\documentclass[preprint,3p,times,review]{elsarticle}

\usepackage{amssymb}
\usepackage{amsmath}
\usepackage{amsthm}
\usepackage{amsfonts}
\usepackage{amssymb}
\usepackage{mathtools}
\usepackage{algorithm,algpseudocode}
\usepackage{enumitem}
\usepackage{soul}
\usepackage{booktabs}

\usepackage{lineno}

\usepackage{xcolor}
\usepackage{hyperref}
\usepackage{tabularray}
\usepackage{subcaption}

\usepackage{calrsfs}
\DeclareMathAlphabet{\pazocal}{OMS}{zplm}{m}{n}

\DeclarePairedDelimiterX{\inner}[2]{\langle}{\rangle}{#1, #2}
\DeclareMathOperator{\Tr}{Tr}
\DeclareMathOperator{\vop}{vec}
\DeclareMathOperator{\Var}{Var}
\newcommand{\R}{\mathbb{R}}
\newcommand{\E}{\mathbb{E}}
\newcommand{\psd}{\mathbb{S}}
\newcommand{\vk}{\mathbf{k}}
\newcommand{\tX}{\pazocal{X}}
\newcommand{\tY}{\pazocal{Y}}

\newcommand{\tA}{\pazocal{A}}
\newcommand{\mX}{\mathbf{X}}

\newcommand{\mR}{\mathbf{R}}
\newcommand{\mK}{\mathbf{K}}
\newcommand{\mA}{\mathbf{A}}
\newcommand{\mU}{\mathbf{U}}
\newcommand{\mV}{\mathbf{V}}
\newcommand{\mM}{\mathbf{M}}

\newcommand{\mI}{\mathbf{I}}
\newcommand{\mC}{\mathbf{C}}
\newcommand{\vx}{\mathbf{x}}
\newcommand{\vy}{\mathbf{y}}
\newcommand{\vr}{\mathbf{r}}
\newcommand{\va}{\mathbf{a}}
\newcommand{\vm}{\mathbf{m}}
\newcommand{\vc}{\mathbf{c}}

\newcommand{\etal}{\textit{et al.}}

\journal{...}

\begin{document}
% \linenumbers
\begin{frontmatter}



\title{ViSTR-GP: Online Cyberattack Detection via \underline{Vi}sion-to-\underline{S}tate \underline{T}ensor \underline{R}egression and \underline{G}aussian \underline{P}rocesses in Automated Robotic Operations}


\author[1]{Navid Aftabi} %% Author name
\ead{aftabi@uw.edu}
\author[2]{Philip Samaha}
\ead{psamaha@clemson.edu}
\author[3]{Jin Ma}
\ead{jin7@clemson.edu}
\author[3]{Long Cheng}
\ead{lcheng2@clemson.edu}
\author[2]{Ramy Harik}
\ead{harik@clemson.edu}

\author[1]{Dan Li\corref{cor1}}
\ead{dli27@uw.edu}
\cortext[cor1]{Corresponding Author: Dan Li, Department of Industrial and Systems Engineering, University of Washington, Seattle, WA 98195, USA (email: \href{mailto:dli27@uw.edu}{dli27@uw.edu})}

\affiliation[1]{
organization={Department of Industrial and Systems Engineering, University of Washington},%Department and Organization
addressline={Box 352650}, 
city={Seattle},
postcode={98195}, 
state={WA},
country={USA}}

\affiliation[2]{
organization={Department of Automotive Engineering, Clemson University},%Department and Organization
addressline={4 Research Dr}, 
city={Greenville},
postcode={29607}, 
state={SC},
country={USA}}

\affiliation[3]{
organization={School of Computing, Clemson University},%Department and Organization
addressline={821 McMillan Rd}, 
city={Clemson},
postcode={29631}, 
state={SC},
country={USA}}

%% Abstract
\begin{abstract}
Industrial robotic systems are central to automating smart manufacturing operations. Connected and automated factories face growing cybersecurity risks that can potentially cause interruptions and damages to physical operations. Among these attacks, data-integrity attacks often involve sophisticated exploitation of vulnerabilities that enable an attacker to access and manipulate the operational data and are hence difficult to detect with only existing intrusion detection or physical model-based detection. This paper addresses the challenges in utilizing existing side-channels to detect data-integrity attacks in robotic manufacturing processes by developing an online detection framework, ViSTR-GP, that cross-checks encoder-reported measurements against a vision-based estimate from an overhead camera outside the controller’s authority. In this framework, a one-time interactive segmentation initializes SAM-Track to generate per-frame masks. A low-rank bilinear tensor-regression surrogate maps each mask to measurements, while a matrix-variate Gaussian process models nominal residuals, capturing temporal structure and cross-joint correlations. A frame-wise test statistic derived from the predictive distribution provides an online detector with interpretable thresholds. We validate the framework on a real-world robotic testbed with synchronized video frame and encoder data, collecting multiple nominal cycles and constructing replay attack scenarios with graded end-effector deviations. Results on the testbed indicate that the proposed framework recovers joint angles accurately and detects data-integrity attacks earlier with more frequent alarms than all baselines. These improvements are most evident in the most subtle attacks. These results show that plants can detect data-integrity attacks by adding an independent physical channel, bypassing the controller's authority, without needing complex instrumentation.
\end{abstract}

\begin{graphicalabstract}
\end{graphicalabstract}

\begin{highlights}
\item Cyberattack detections on industrial robots via cross-checking of side-channel and network data.
\item Mapping video frames to measurements using a compact learned model.
\item Statistical modeling of normal behavior enables reliable, real-time alarms.
\item Detecting subtle replay attacks on a real-world testbed.
\end{highlights}

%% Keywords

\end{frontmatter}

\section{Introduction}\label{section:1}
The progress in digital transformation, real-time data analysis, and Artificial Intelligence (AI) is driving manufacturing towards integrated Industry~4.0 environments. Consequently, cyberattacks targeting physical infrastructure, particularly in smart manufacturing systems, are increasingly prevalent~\cite{TUPTUK201893}. Recently, manufacturing systems have been one of the most targeted sectors for cyberattacks~\cite{DESMIT2017339}. Recent evidence from manufacturing underscores the rise of threats and the limits of cyber-only defenses~\cite{RAHMAN2022676}. Stuxnet~\cite{Farwell01022011} stands out as one of the most infamous cyberattacks targeting a physical system. Stuxnet achieved its aims by stealthily altering programmable logic controller (PLC) commands while feeding a loop of normal sensor readings back to operators. Other notable incidents like the 2014 German steel-mill breach \cite{lee2014german}, WannaCry shutdowns at auto plants \cite{williams2022taxonomy,9247392}, and the 2019 LockerGoga attack on Norsk Hydro \cite{9247392} highlight the potential for cyberattacks to inflict physical damage on manufacturing systems.

As the core of smart manufacturing, industrial robotic systems have come under the spotlight as critical and vulnerable cyber-physical assets. Modern factory robots are increasingly connected (for remote monitoring, programming, and integration into smart production lines), which opens avenues for cyberattacks to compromise their controllers and could have a profound impact if compromised~\cite{YANG2025104586}. In 2017~\cite{Quarta7958582}, researchers showed that hacking an industrial robot (such as ABB IRB140) could cause defects, halt production, and pose risks to machinery and humans by altering its trajectory. They also demonstrated how the robot's interface and logs could be manipulated to falsely reassure operators of normal operations. Recent research has outlined that data-integrity attacks on industrial robots can cause significant damage, misinformation, or operational disruptions by compromising the data accuracy, completeness, or reliability of data~\cite{Pu9839629}. Replay attacks~\cite{Pu3430775,PU2024186,Yolaçan10855442} and false-data injection attacks~\cite{app12062765,Santoso9812734} are the primary cyberattacks studied in the literature. These attacks underscore the real danger of undetected cyber-physical attacks: an intruder with access to a robot’s sensory or control network can subtly manipulate motions or replay historical sensor readings, leading to hidden defects and safety hazards that traditional monitoring might miss.

Current methods for detecting cyberattacks on industrial robots span model-based estimation~\cite{Alemzadeh7579758,Hector9473818,Longari10508318}, data-driven inference~\cite{Yolaçan10855442,app12062765,Santoso9812734,Narayanan3264894,LI2024340,shao2024covert,ORABI2024591}, and formal reasoning over controller or PLC commands~\cite{Zhang8835244,Sun9581186}, with some works validating network data against side-channels~\cite{YANG2025104586,Pu3430775,Pu9355028,SONG2023376}. In practice, these approaches face recurring challenges: (i) detectors that reason solely over embedded measurements can be manipulated or replayed, leaving sophisticated data-integrity attacks undetected; (ii) physics- and formal-methods demand high-fidelity models or properties that are brittle and costly to maintain across multi-joint nonlinear motions; (iii) learning-based detectors trained with labeled data often generalize poorly to unseen manipulations; and (iv) side-channel defenses, while promising, require extra hardware and can be sensitive to disruptions and noise~\cite{Pu9839629}. Motivated by these challenges, we propose a data-driven online detection framework that: (i) uses an independent physical viewpoint via an overhead camera outside the controller’s authority to estimate joint motion that shows the ground-truth about the robotic system; (ii) learns a lightweight, frame-to-state surrogate using only the nominal data, reducing dependence on labeled attacks; and (iii) requires only a fixed camera and no additional side-channel instrumentation. Our framework employs a sequential cross-view test with interpretable, per-frame thresholds to raise timely alarms at controlled false-alarm rates. Vision-based authentication in additive manufacturing supports this strategy by detecting cyber-induced path changes from video alone~\cite{MAMUN2022429}.

\subsection{Related Work}\label{section:1.1}
Industrial robots are vulnerable to cyber-physical attacks due to their connectivity and integration with physical processes. Attackers can directly manipulate operations by uploading malicious programs, injecting motion commands, or altering controller settings (e.g., PID gains, safety limits), causing trajectory errors and product defects while logs appear normal~\cite{Pu9839629}. Data deception is dangerous, as attackers modify or replay real-time data during operations, intercepting and altering transmitted data to conceal the manipulated physical trajectory from detectors monitoring authentic encoder streams~\cite{PU2024186}. Denial-of-service (DoS) attacks hinder production by overloading controller communication, locking robots, or encrypting files or parameters~\cite{Pu9839629}. Sensors can be exploited through firmware calibration tampering or physical signal injection at specific frequencies. This exploitation corrupts feedback and bypasses traditional detection mechanisms, thus propagating faults in the control loop~\cite{Shaik70023}. Attackers aim to subtly degrade quality or cause equipment damage and safety hazards~\cite{PU2024186}. Notably, experimental studies show that altering PID parameters after controller compromise significantly affects accuracy, showing how minor integrity breaches cause physical misalignment unnoticed by standard monitoring systems~\cite{Quarta7958582}. These patterns highlight the need for defenses that verify robot behavior through independent physical evidence instead of relying solely on network data.

Existing cyberattack detection frameworks for industrial robots mainly fall into four main categories: physics-based, formal, learning-based, and side-channel~\cite{Pu9839629}. Physics-based methods predict motion using dynamics or kinematics models and alert on significant deviations from actual paths. They avoid hardware alterations but struggle with multi-joint nonlinear systems~\cite{Pu9839629}. These methods can be tricked by replaying legitimate data and tend to generate frequent false alarms~\cite{Alemzadeh7579758,Hector9473818,Longari10508318}. Formal methods define normal robot behavior through safety properties from PLC or robot code, I/O mappings, and operational data. Model checking identifies property violations but overlooks replay and covert attacks where process changes and measurement replays make properties seem satisfied. Model checking is computationally demanding, and aligning software with physical dynamics to create comprehensive cyber-physical properties is difficult~\cite{Zhang8835244,Sun9581186}. Side-channel methods verify robot data by comparing it with independent physical signals, such as energy consumption or motion. These signals are isolated from the production network, making them difficult for attackers to spoof. These methods are promising but require extra hardware and are sensitive to disruptions and noise~\cite{YANG2025104586,Pu3430775,Pu9355028}.

Learning-based approaches conceptualize cyberattack detection as a data-driven classification task over observed robotic measurements. During the training phase, these methods collect both normal and abnormal trajectories. They develop a parameterized data-driven model using supervised or unsupervised techniques to distinguish normal from abnormal behaviors. Subsequently, the model is deployed on real-time data~\cite{Yolaçan10855442,app12062765,Santoso9812734,Narayanan3264894,ORABI2024591}. Given that these models are developed using labeled data of normal and abnormal behaviors, their capacity to identify anomalous signals is restricted to those observed during the training phase. This limitation hampers their ability to generalize to unseen abnormal behaviors. Recently, hybrid learning-physics-based approaches~\cite{LI2024340,shao2024covert} and learning-side-channel-based methodologies~\cite{app12062765,SONG2023376} have been introduced to complement each other's capabilities, showing considerable promise. Our approach is classified under learning-side-channel-based methodologies, utilizing an overhead camera as a side-channel. It is developed exclusively using nominal operation data, enabling the straightforward detection of anomalies in robotic operations.

With the rapid advancement of computer vision techniques and deep learning architectures, automated visual analysis has emerged as a powerful approach for detecting anomalies in manufacturing environments. Modern computer vision systems can extract rich and complementary cues from images and video streams on production lines such as texture, geometry, motion, and temporal patterns, enabling accurate and efficient detection of deviations from normal operations. A major part of existing work focuses on detecting anomalies in the final product. Earlier industrial vision systems similarly combine classical geometry with lightweight CNNs for fast, reliable product inspection on lines~\cite{WANG201952}. Recently, Roth~\etal~\cite{roth2022towards} proposed PatchCore, which leverages pre-trained Convolutional Neural Network (CNN) features and a memory bank of normal patch embeddings to identify visual outliers with state-of-the-art accuracy. You~\etal~\cite{you2022unified} introduced UniAD, a unified anomaly detection model capable of handling multiple object classes without requiring a separate model per class. Li~\etal~\cite{li2025multi} further extended this line of work with MulSen-AD, a multi-sensor framework that fuses RGB, infrared, and 3D data to detect diverse types of defects not visible in single-modality inputs. In contrast to these product-focused approaches, this paper targets the detection of anomalous \textit{robot behavior} during manufacturing processes. Schirmer~\etal~\cite{schirmer2023anomaly} employed an LSTM-based autoencoder to detect deviations in human-robot collaborative assembly sequences, distinguishing between benign and critical anomalies using a directed task graph. Okazaki~\etal~\cite{okazaki5298327spatio} proposed a spatio-temporal video analysis method that aligns full robot assembly cycles using dynamic time warping and identifies anomalies by comparing motion patterns to a learned codebook. Inceoglu~\etal~\cite{inceoglu2023multimodal} developed FINO-Net, a deep model that fuses RGB-D and exteroceptive sensor data to detect and classify robot manipulation failures in real time, without the need for manually annotated fault types. Collectively, these approaches highlight the promise of vision analysis for recognizing abnormal robotic operations in complex, real-world manufacturing settings. 

\subsection{Challenges and Contributions}\label{section:1.2}
Building on the foregoing discussion, we distill four recurring challenges in cyberattack detection for industrial robots: 
\begin{enumerate}[label=(\roman*)]
% [label=\textbf{G\arabic*:},wide,labelwidth=!, labelindent=0pt]
\item Methods that reason solely over embedded sensor data from the robot are vulnerable to the replay and other covert-type attacks that have access or knowledge of robot's normal behavior. In these attacks, the adversary can craft benign‐looking streams while physically deviating the motion.
\item Some commonly used Side-channels such as power consumption monitoring requires intrusive installation and are sensitive to disruption and noises.
\item Even though the high-fidelity dynamic or kinematic representations provide the accurate modeling of the system behavior, these parametric models often rely on detailed specifications and nonlinear models. The nonlinearity poses computational challenges in detection methods and lacks generality across different systems. 
\end{enumerate}
We present an online data-driven framework to address these challenges by inferring the robot's joint states from video in real time, comparing them to observed joint angles, and using sequential detection to reveal cyberattacks while controlling false alarms. The key contributions of this paper are as follows:
\begin{enumerate}[label=(\roman*)]
    % problem formulation with an independent physical reference as ground-truth
    \item We formalize cyberattack detection in industrial robotic systems as a cross-view residual test. This test compares encoder-reported joint angles with vision-based estimated angles from an overhead camera that is outside the controller’s authority. This explicitly targets data-integrity attacks that keep internal measurements within nominal bounds. The framework requires no additional plant instrumentation beyond a fixed camera, no labeled data for training, and minimal one-time interaction for segmentation.
    \item We use a fixed overhead camera as an independent side-channel to illustrate the ground-truth of the physical channel, decoupled from system-based noises and disruptions. The SAM-Track generates clear frame-by-frame foreground masks, reducing background and illumination noise to stabilize the visual signal. Our framework is developed only using nominal cycles that effectively scale well to any unseen anomaly by cross-viewing vision-estimated and encoder-reported joint states.
    \item ViSTR-GP is a model-free framework. Instead of high-fidelity models, we adopt a data-driven pipeline. A low-rank bilinear tensor-regression (TR) map directly converts masks to joint angles, efficiently capturing the (nonlinear) image-to-state relationship with few parameters and strong generalization capability. Nominal residuals are then modeled by a matrix-variate Gaussian process (MVGP), capturing temporal and cross-joint correlations for time-adaptive uncertainty and interpretable frame-wise decision rules. This avoids costly physics identification, remains computationally lightweight for online use, and is portable across tasks with minimal recalibration.
    % % vision–to–state estimator (TR)
    % \item We develop an efficient state surrogate that maps per-frame masks onto joint angles via a low-rank bilinear tensor-regression (TR) model. This method provides a fast, label-efficient estimator with the capacity to generalize over cycles. 
    
    % We use camera that is less sensitive to noise, we traine on the normal data, we use masks SAM-Track that reduces noises. Cross-view of ground-truth and sensor data.
    % \item Nonlinearity. Talk about TR and GP that are data-driven and not model or physics base.
    % % time-dependent residual modeling (MVGP)
    % \item We model nominal residuals with a matrix-variate Gaussian process (MVGP) that captures temporal structure within a cycle and cross-joint correlations. A frame-wise $\chi^2$ test derived from the MVGP predictive distribution delivers an online detector with interpretable thresholds and controllable false-alarm rates.
    % real-world experiments
    \item The effectiveness of the proposed framework is validated using a real-world physical testbed. In this setup, a synchronized camera–encoder dataset is created on a robotic assembly line. Replay-attack scenarios with graded, sub-centimeter end-effector deviations are constructed. Our method is benchmarked against three representative baselines, demonstrating superior rapid detection under the most subtle replay attacks.
\end{enumerate}

The remainder of this paper is organized as follows: Section~\ref{section:2} covers preliminaries and notation; Section~\ref{section:3} details the formalization of the detection problem; Section~\ref{section:4} outlines the methodology, covering the vision-based state estimator, residual modeling, and the online decision rule; Section~\ref{section:5} details the robotic work-cell, hardware, data acquisition, synchronization, and attack scenarios; Section~\ref{section:6} presents experimental results and baselines; Section~\ref{section:7} discusses key findings, implications, limitations, and future extensions; and Section~\ref{section:8} concludes the paper.

\section{Preliminaries}\label{section:2}
In Section~\ref{section:2.1}, the notation and basics in multi-linear algebra used in this paper are reviewed. Afterwards, the methods used in the detection pipeline are briefly reviewed in Sections~\ref{section:2.2} and~\ref{section:2.3}. 
% Then, the cyberattack detection problem is formally defined in Section~\ref{section:2.3}.  % update this

\subsection{Notation and Multi-linear Algebraic Operations}\label{section:2.1}
In preparation for the ensuing discourse, we first introduce the notation employed within this paper. Tensors are represented using calligraphic letters, for instance, $\tX$; matrices are depicted by boldface uppercase letters, such as $\mX$; vectors are illustrated with boldface lowercase letters, exemplified by $\vx$; and scalars are denoted by lowercase letters, e.g., $x$. The order of a tensor denotes the number of its dimensions, also referred to as ways or modes. The element $(i_1,i_2,\dots,i_N)$ of a tensor of order $N$, denoted by $\tX$, is specified by $\tX(i_1,i_2,\dots,i_N)$, with indices conventionally extending from their lowercase to uppercase counterparts, e.g., $i_n\in\{1,2,\dots,I_n\}$. The vectorization operator is denoted as $\vop(\cdot)$. The inner product for two tensors of identical size, denoted as $\tX$ and $\tY$, is specified by $\inner{\tX}{\tY}= \sum_{i_1}\dots\sum_{i_N}\tX(i_1,i_2,\dots,i_N)\tY(i_1,i_2,\dots,i_N)$. Furthermore, the Frobenius norm is articulated as $\Vert\tX\Vert_{F} = \sqrt{\inner{\tX}{\tX}}$. The multiplication of a tensor $\tX$ by a matrix $\mU\in\R^{J_n\times I_n}$ along mode $n$ yields a resultant tensor of size $I_1\times \dots \times I_{n-1}\times J_n \times I_{n+1} \times \dots \times I_N$. This operation involves the multiplication of every vector fiber in mode $-n$ with $\mU$, represented as $(\tX\times_{n}\mU)(i_1\times \dots \times i_{n-1}\times j_n \times i_{n+1} \times \dots \times i_N)= \sum_{i_n} \tX(i_1,\dots,i_N) \mU(j_n,i_n)$. Analogously, the multiplication of a tensor $\tX$ by a vector $\vy\in\R^{I_n}$ along mode $n$ results in a tensor of size $I_1\times \dots \times I_{n-1}\times I_{n+1} \times \dots \times I_N$, corresponding to $(\tX\overline{\times}_n\vy)(i_1\times \dots \times i_{n-1}\times i_{n+1} \times \dots \times i_N)= \sum_{i_n} \tX(i_1,\dots,i_N) \vy(i_n)$. The mode\(-n\) unfolding (or matricization) of a tensor \(\tX\in\R^{I_1\times\cdots\times I_N}\) rearranges its entries into the matrix \(\mX_{(n)}\in\R^{I_n\times \prod_{m\neq n} I_m}\) whose columns are the mode\(-n\) fibres of \(\tX\). We denote this operation by \(\tX_{[n]}\). We write $\mI_d$ for the $d\times d$ identity matrix, $\mathbf{0}_{m\times n}$ for the $m\times n$ zero matrix, and $\mathbf{0}_d$ for the $d$–dimensional zero vector.  For any matrix $\mX\in\R^{m\times n}$, $\mX^\top$ denotes its transpose, $\Tr(\mX)=\sum_{i=1}^{\min\{m,n\}}x_{ii}$ its trace, and $\mX^{\dagger}$ its Moore–Penrose pseudoinverse.

\subsection{Tensor Decomposition}\label{section:2.2}

% \subsubsection{Tensor Decomposition}\label{section:2.3.2}
% Video frames collected during one production cycle constitute a third-order data tensor \(\tX\in\R^{T\times H\times W}\). Engaging in direct learning \(\pazocal{F}_{\theta}:\R^{H\times W}\rightarrow\R^{J}\) on raw pixel data proves to be statistically inefficient and computationally demanding, primarily due to the exponential growth in the number of free parameters with \(H\cdot W\). 

Techniques involving tensor decomposition effectively disentangle spatial and temporal variations, producing a low-dimensional representation. This representation not only expedites the learning process but also enhances generalization capabilities when handling moderate volumes of data. Tucker decomposition~\cite{tucker} is widely utilized as one of the predominant methods for tensor decomposition, esteemed for its capacity as a higher-order extension of the matrix singular value decomposition (SVD). Consider $\tX\in\R^{I_1\times\dots\times I_N}$ as an $N^{th}-$order tensor, whereupon the Tucker model is formulated accordingly. 
\[
\tX = \pazocal{G}\times_1\mU^{(1)}\times_2 \mU^{(2)}\times_3 \dots\times_N\mU^{(N)},
\]
Here, $\pazocal{G}\in\R^{R_1\times\dots\times R_N}$ represents the core tensor, while $\mU^{(n)}\in\R^{R_n\times I_n}$ signifies the mode$-n$ factor matrix. The core matrix encapsulates the interactions among various modes, and frequently, owing to \(R_n \le I_n\), the core tensor is regarded as the compressed version of the original tensor. If all factor matrices $\{\mU^{(n)}\}_{n=1}^{n=N}$ possess column-wise orthonormality and the core tensor $\pazocal{G}$ exhibits all-orthogonality (meaning any sub-tensors are orthogonal) and is also ordered, this decomposition is classified as a higher-order singular value decomposition (HOSVD). An inherent method for determining the components of the Tucker decomposition involves minimizing the squared error between the original tensor and its corresponding reconstruction, which is
\begin{equation}\label{eq:tucker}
    \left\{\pazocal{G}, \mU^{(1)}, \dots, \mU^{(N)}\right\} \in \arg\min_{\pazocal{G}, \mU^{(1)}, \dots, \mU^{(N)}} \left\Vert \tX - \pazocal{G}\times_1\mU^{(1)}\times_2 \mU^{(2)} \dots\times_N\mU^{(N)}  \right\Vert^2.
\end{equation}

\subsection{Matrix-variate Gaussian Processes}\label{section:2.3}
A random matrix $\mX\in\R^{n\times m}$ is considered to possess a matrix-variate Gaussian distribution with a mean matrix $\mM\in\R^{n\times m}$ and covariance matrices $\mV\in\psd_+^n$, $\mU\in\psd_+^m$ if and only if its probability density function is articulated as
\[
p(\mX|\mM, \mV,\mU)= \frac{\exp\left(-\frac{1}{2}\Tr\left(\mU^{-1}(\mX - \mM)^\top \mV^{-1} (\mX - \mM)\right)\right)}{(2\pi)^{\frac{nm}{2}} \det(\mV)^{\frac{m}{2}} \det(\mU)^{\frac{n}{2}}}
\]
such that $\psd_+^n$ and $\psd_+^m$ represent the set of symmetric positive semi-definite (PSD) matrices~\cite{chen2020multivariate}. This distribution is represented as $\mX\sim\pazocal{MN}_{n\times m}(\mM,\mV,\mU)$. For any matrix-variate Gaussian distribution, the relationship $\mX^\top\sim\pazocal{MN}_{m\times n}(\mM^\top,\mU,\mV)$ holds. Furthermore, $\mX\sim\pazocal{MN}_{n\times m}(\mM,\mV,\mU)$ is valid if and only if $\vop(\mX)\sim\pazocal{N}\left(\vop(\mM), \mU\otimes\mV\right)$, where $\otimes$ signifies the Kronecker product.

A matrix-variate Gaussian process (MVGP) constitutes a stochastic process whereby, for any finite collection of locations, the joint distribution of the process's values at those points is characterized by a matrix-variate normal distribution~\cite{yan2012sparsematrixvariategaussianprocess}. Consequently, the output is represented not as a vector but as a matrix of random variables, with the interrelationships among these variables being encapsulated through the properties intrinsic to the matrix-variate normal distribution. Let $\mathbf{f}$ be a multivariate Gaussian process on $\mathcal{T}$ with vector-valued mean function $\mu:\mathcal{T}\to\R^m$ with covariance (kernel) function $k:\mathcal{T}\times\mathcal{T}\to\R$ and positive semi-definite parameter matrix $\Sigma\in\psd_+^m$. Then, any finite collection of vector-valued variables has a joint matrix-variate Gaussian distribution:
\[
\left[\mathbf{f}(t_1)^\top,\mathbf{f}(t_2)^\top,\dots,\mathbf{f}(t_n)^\top\right]^\top \sim \pazocal{MN}_{n\times m}(\mM,\mK,\Sigma), \qquad t_i\in\mathcal{T},
\]
where \(\mM\in\R^{n\times m}\) with \([\mM]_{ij}= \mu_j(t_i)\), $i=1,\dots,n$, $j=1,\dots,m$, and \(\mK\in\R^{n\times n}\) with \([\mK]_{ij} = k(t_i,t_j)\) and $i,j=1,\dots,n$. Then, we denote $\mathbf{f} \sim \pazocal{MGP}(\mu, k, \Sigma).$

\section{Problem Formulation}\label{section:3}
% \begin{itemize}
%     \item System Dynamics and Data Streams
%     \item Adversarial Threat Model
%     \item Attack Detection
% \end{itemize}

We consider an automated robotic work-cell which executes a predetermined task in a repetitive manner. During each cycle, two synchronized data streams are accessible. First, the frames of the video are acquired through an overhead camera. A frame captured at time instance $t$ is represented as $\mX_{t}\in\R^{H\times W}$, where $H$ and $W$ correspond to the height and width of the frame, respectively. Second, the encoder-reported measurement signals are analyzed within the context of the robot's joint angles. At a given time $t$, the measurement signals are represented as $\va_t\in\R^J$, where $J$ indicates the number of joints present in the robot within the automated production line. These signals are utilized by the controller to ascertain subsequent control actions. 

The duration of the task (cycle) is defined by \(T\) time steps, contingent upon the internal communication buses and the camera frame rate, which correspond to the number of observed frames and signals. The camera is assumed tamper-proof, whereas networked motion controllers and internal communication buses are potentially exposed to cyberattacks.

Let \(\va^A\in\R^{J}\) represent the joint-angle executed by the robot, which may be subject to compromise. A data-integrity attack can be executed by injecting additive perturbations $\va^A_t = \va_t + \delta_t,$ $\delta_t\in\R^{J},$ within constraints of actuator limits, timing, or stealthiness. Another common attack is a replay attack that does not require complete understanding of robotic dynamics. In this scenario, the attacker collects a series of measurements $\{\va_t^r\}_{t=1}^{t=T}$ under normal operating conditions. Once the attack is initiated, the attacker substitutes the current accurate measurement signal $\va_t$ with previously recorded measurements $\va_t^A=\va_t^r=\va_{t-\Delta t}$ to mislead the detection mechanisms while simultaneously altering the control actions. 

Given the exclusion of the camera from the control authority, the integrity of the video stream \(\{\mX_t\}_{t=1}^{t=T}\) remains intact. Consequently, any non-zero \(\delta_t\) or any attack actions that elude the detector will ultimately result in a physical discrepancy between the joint angles inferred from visual input and the signals provided by the encoders. Inspired by this rationale, it is essential to establish a mapping \(\pazocal{F}_{\theta}: \R^{H\times W}\to\R^{J}\) capable of estimating joint angles based on a singular frame and associated model parameters \(\theta\). Applying \(\pazocal{F}_{\theta}\) across each frame results in the generation of predicted angles $\widehat{\va}_t = \pazocal{F}_\theta(\mX_t)$, and the residual vector $\vr_t=\va_t - \widehat{\va}_t$ for $t=1,\dots,T$. Under normal operation conditions, the residuals follow a stochastic law
\(\vr_t \sim \mathcal{M}_0\) whose first- and second-order properties can be learned offline.  
Cyber manipulation manifests as a change in distribution:
\[
    \mathcal{H}_0:  \vr_t\sim\mathcal{M}_0
    \quad\text{vs.}\quad
    \mathcal{H}_1:  \exists t^\ast\le t\le T : \vr_t\sim\mathcal{M}_1 \neq \mathcal{M}_0.
\]
The objective is to design a sequential decision rule that raises an alert at time \(\tau_{\text{alarm}}\in\{1,\dots,T\}\cup\{\infty\}\), ensuring that (i) \(\Pr(\tau_{\text{alarm}}=\infty\mid\mathcal{H}_0)\le\alpha\) is maintained (controlled false-alarm rate) and (ii) minimizes the detection delay defined by \(\E[\tau_{\text{alarm}}-t^{\ast}\mid\mathcal{H}_1]\) (rapid detection).  

% The remainder of Section~\ref{section:2} reviews the research background that supplies the ingredients \((\pazocal{F}_{\theta},\mathcal{M}_0)\) for the proposed online detection framework.

\section{Methodology}\label{section:4}
\begin{figure}[!ht]
    \centering
    \includegraphics[width=.8\linewidth,page=3]{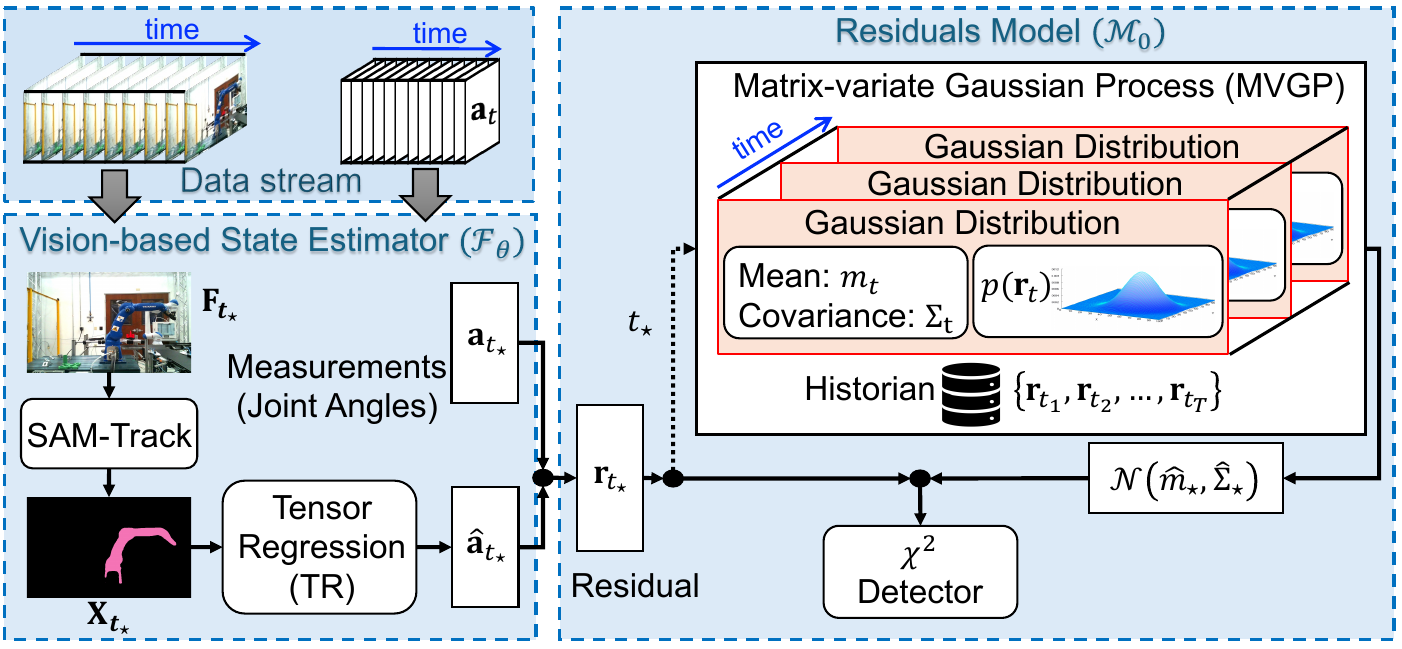}
    \caption{Overall online cyberattack detection framework.}
    \label{fig:fw}
\end{figure}

Figure~\ref{fig:fw} illustrates an overview of the proposed detection pipeline and clarifies how the two key building blocks introduced in Section~\ref{section:3} interact in real time.  The left pane denotes the vision-based state estimator~\(\pazocal{F}_{\theta}\).  Each incoming camera frame is first segmented and tracked by SAM-Track, producing a mask \(\mX_t\) that isolates the robot body from background clutter. A tensor-regression (TR) model then maps \(\mX_t\) to a point estimate \(\widehat{\va}_t\in\R^{J}\) of the joint-angle vector, thereby providing an vision-driven surrogate for the encoder readings \(\va_t\). The difference \(\vr_t=\va_t-\widehat{\va}_t\) constitutes a low-dimensional, physically interpretable residual signal that is agnostic to controller details and therefore robust to cyberattacks.

The right pane encapsulates the residuals model~\(\mathcal{M}_{0}\).  Residuals produced under normal operation are stored in an on-board historian and used to train a matrix-variate Gaussian process (MVGP), yielding a predictive mean \(\widehat{\vm}_{t_\star}\) and covariance \(\widehat{\Sigma}_{t_\star}\) for the forthcoming residual at time \(t_\star\). In real-time operations, each newly observed \(\vr_t\) is assessed against its predictive distribution utilizing the Mahalanobis distance. Assuming the null hypothesis ($\mathcal{H}_0$), this distance characteristically adheres to a \(\chi^{2}\) distribution with \(J\) degrees of freedom. A sequential threshold test therefore offers a principled way to raise an alarm once the distributional shift \(\mathcal{M}_{0}\to\mathcal{M}_{1}\) becomes statistically significant. The remainder of this section provides the technical details of (i) learning \(\pazocal{F}_{\theta}\) via tensor regression (Section~\ref{section:4.1}), (ii) specifying and estimating the MVGP parameters (Section~\ref{section:4.2}), and (iii) deriving the real-time \(\chi^{2}\) detector and its operating characteristics (Section~\ref{section:4.3}).

\subsection{Vision-Based State Estimation}\label{section:4.1}

\begin{figure}[!ht]
    \centering
    \includegraphics[width=\linewidth, page=2]{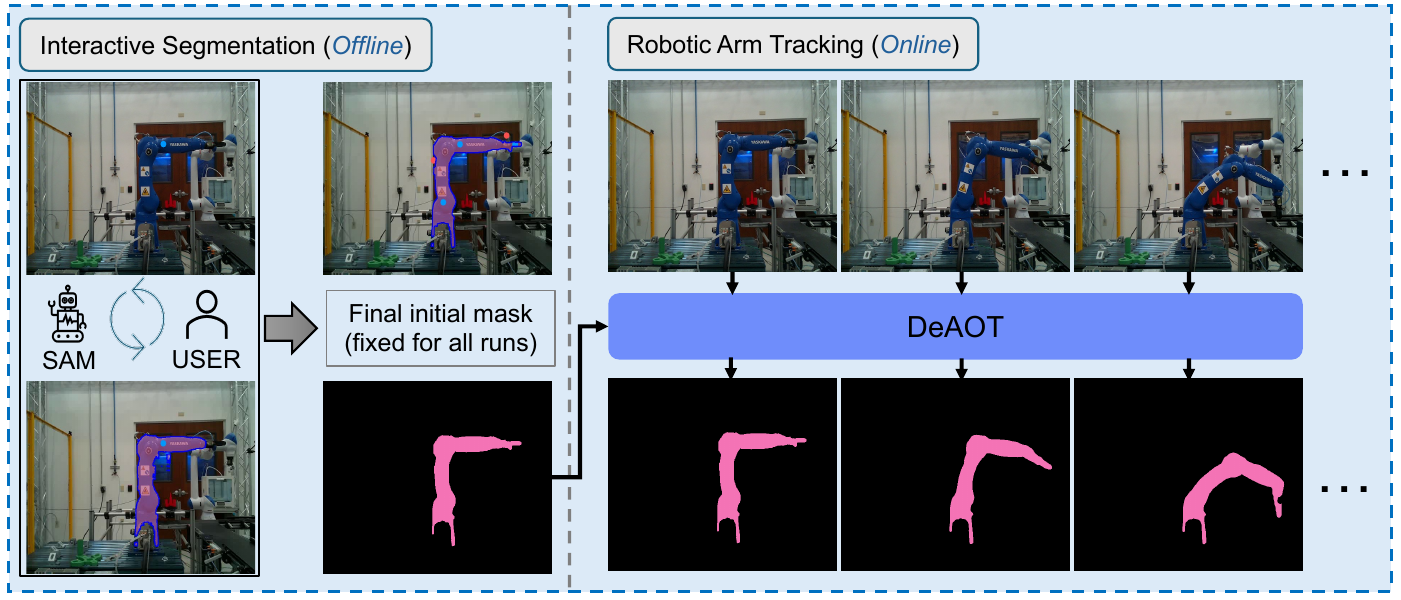}
    \caption{The pipeline of our frame-to-mask process, with an offline interactive segmentation stage and online robotic arm tracking stage.}
    \label{fig:sat_pipeline}
\end{figure}

A foreground mask is first extracted from each raw frame by the SAM-Track pipeline. SAM-Track~\cite{cheng2023segment} couples a one-time interactive first-frame segmentation via Segment Anything (SAM)~\cite{kirillov2023segany} with a Decoupling Features in Hierarchical Propagation (DeAOT) tracker~\cite{yang2022decoupling} that propagates the mask across the video. In our work-cell, the camera is fixed and the arm returns to a home pose each cycle, so a single user-verified mask (e.g., a few positive or negative clicks) is saved and reused for all online runs, eliminating further manual input. SAM-Track then produces a time-ordered sequence of foreground masks that isolates the robot body and is robust to background and illumination clutter. We treat the tracker as plug-and-play with no additional tuning. The pipeline is shown in Fig.~\ref{fig:sat_pipeline}.

The SAM-Track isolates the robot body and yields a matrix $\mX\in\R^{H\times W}$ that is largely invariant to background clutter and illumination. Conditional on this mask, the vision–based estimator $\pazocal{F}_\theta$ is postulated to be a bilinear map of the form
\begin{equation}
    \pazocal{F}_\theta(\mX_t) = \pazocal{B}_h \mX_t \pazocal{B}_w + \varepsilon, \qquad \theta=\{\pazocal{B}_h, \pazocal{B}_w\},
\end{equation}
where $\varepsilon$ is the modeling error and $\pazocal{B}_h\in\R^{J\times H}$ and $\pazocal{B}_w\in\R^{W}$ are basis matrices that compress spatial information along the image height ($H$) and width ($W$), respectively, and jointly produce the $J-$dimensional joint-angle estimate $\widehat{\va}_t$.

In order to estimate $\pazocal{B}_h$ and $\pazocal{B}_w$, we shall define a third-order tensor $\tX\in\R^{T\times H\times W}$. This tensor is formed through the concatenation of the masked-frame stream \(\{\mX_t\}_{t=1}^{t=T}\). We also define $\mA\in\R^{T\times J}$ through the concatenation of the joint-angles stream $\{\va_t\}_{t=1}^{t=T}$. This pertains to an operation with a cycle-length of $T$, wherein each $\va_t$ is associated with its respective $\mX_t$. Assuming that $N$ replications of the identical operations, $\tX_1, \dots, \tX_N$ and $\mA_1, \dots,\mA_N$, have been collected, the problem thus is to solve
\begin{equation}\label{eq:tr_problem}
    \left\{\widehat{\pazocal{B}}_h, \widehat{\pazocal{B}}_w \right\} \in \arg\min_{\pazocal{B}_h,\pazocal{B}_w} \frac{1}{2} \sum_{i=1}^{N} \left\Vert\mA_i - \tX_i \times_2 \pazocal{B}_h \overline{\times}_3 \pazocal{B}_w\right\Vert_F^2
\end{equation}

To enhance the efficiency in solving Problem~\eqref{eq:tr_problem}, we employ Tucker decomposition on each $\tX_i$, specifically $\tX_i \approx \pazocal{G}_i \times_1 \mU_t\times_2 \mU_h \times_3 \mU_w$, where $\pazocal{G}_i\in\R^{R\times P\times Q}$ serves as the core tensor derived from the $i^{\text{th}}$ sample, while $\mU_t\in\R^{T\times R}$, $\mU_h\in\R^{H\times P}$, and $\mU_w\in\R^{W\times Q}$ constitute the learned basis matrices. The core tensors $\pazocal{G}_i$ alongside the basis matrices $\mU_t$, $\mU_h$, and $\mU_w$ are obtained by solving Problem~\eqref{eq:tr_tucker}.
\begin{equation}\label{eq:tr_tucker}
    \left\{\pazocal{G}_i, \mU_t, \mU_h, \mU_w\right\} \in \arg\min_{\pazocal{G}_i, \mU_t, \mU_h, \mU_w} \sum_{i=1}^{N} \left\Vert\tX_i - \pazocal{G}_i \times_1 \mU_t \times_2 \mU_h \times_3 \mU_w  \right\Vert_F^2
\end{equation}

The samples can be concatenated to form a new higher-order tensor $\tY\in\R^{N \times T\times H\times W }$. According to the definition of the Frobenius norm, it holds that $\Vert\tY\Vert_F^2 = \sum_{i=1}^{N} \Vert \tX_i \Vert_F^2$. Consequently, the solution to the Problem~\eqref{eq:tr_tucker} is equivalent to that of Problem~\eqref{eq:tr_tucker2} and $\pazocal{G}_i=\left(\pazocal{S} \times_1 \mU_n\right)(i,:,:,:)$, wherein $\mU_n\in\R^{N\times D}$ and $\pazocal{S}\in\R^{D\times R\times P \times Q}$.
\begin{equation}\label{eq:tr_tucker2}
    \left\{\pazocal{S}, \mU_n, \mU_t, \mU_h, \mU_w\right\} \in \arg\min_{\pazocal{S}, \mU_n, \mU_t, \mU_h, \mU_w} \left\Vert \tY - \pazocal{S} \times_1 \mU_n \times_2 \mU_t \times_3 \mU_h \times_4 \mU_w  \right\Vert_F^2
\end{equation}

To solve Problem~\eqref{eq:tr_problem}, we employ the identical rationale, concatenating the measurement matrices to form a higher-order tensor $\tA\in\R^{N\times T\times J}$, while defining $\pazocal{V}=\pazocal{S} \times_1 \mU_n \times_2 \mU_t $, $\pazocal{V}\in\R^{N\times T\times P \times Q}$, $\pazocal{C}_h = \pazocal{B}_h \mU_h $, and $\pazocal{C}_w = \mU_w^\top \pazocal{B}_w $. Subsequently, we solve Problem~\eqref{eq:tr_problem1} with respect to $\pazocal{C}_h$ and $\pazocal{C}_w$ by utilizing the Alternating Least Squares (ALS) algorithm~\cite{Leeuw1994}. Then, we compute $\pazocal{B}_h = \pazocal{C}_h \mU_h^\dagger $, and $\pazocal{B}_w = {\mU_w^\dagger}^\top \pazocal{C}_w $. Algorithm~\ref{alg:ALS} provides the pseudocode for the estimation of $\pazocal{B}_h$ and $\pazocal{B}_w$.
\begin{equation}\label{eq:tr_problem1}
    \left\{\widehat{\pazocal{C}}_h, \widehat{\pazocal{C}}_w \right\} \in \arg\min_{\pazocal{C}_h,\pazocal{C}_w} \frac{1}{2} \left\Vert \tA - \pazocal{V} \times_3 \pazocal{C}_h \overline{\times}_4 \pazocal{C}_w\right\Vert_F^2
\end{equation}

\begin{algorithm}[h]
\caption{Estimation of \(\pazocal{B}_{h}\) and \(\pazocal{B}_{w}\) via Tucker Decomposition \& ALS}
\label{alg:ALS}
\hspace*{\algorithmicindent} \textbf{Input:} \(\tA\in\R^{N\times T\times J}\), \(\tY\in\R^{N\times T\times H\times W}\); 
ranks \(P,Q\); tolerance \(\varepsilon\). \\
\hspace*{\algorithmicindent} \textbf{Output:} Estimated parameters \(\widehat{\pazocal{B}}_{h}\in\R^{J\times H}\), \(\widehat{\pazocal{B}}_{w}\in\R^{W}\).
\begin{algorithmic}[1]

\State \textbf{[Tucker compression]} Compute Tucker decomposition of \(\tY\)\\
\hspace*{6mm}\(\tY\approx\pazocal{S}\times_{1}\mU_{n}\times_{2}\mU_{t}\times_{3}\mU_{h}\times_{4}\mU_{w}\)  
with \(\mU_{h}\in\R^{H\times P}, \mU_{w}\in\R^{W\times Q}\).\vspace{2pt}

\State Form \(\pazocal{V}\gets\pazocal{S}\times_{1}\mU_{n}\times_{2}\mU_{t}\in\R^{N\times T\times P\times Q}\).

\State Initialize \(\pazocal{C}_{h}^{(0)}\in\R^{J\times P}\) and \(\pazocal{C}_{w}^{(0)}\in\R^{Q}\) entries.

\While{not converged}
    \State \textbf{Update \(\pazocal{C}_{h}\):} Form $\tilde{\mV}_k:=\left(\pazocal{V}\times_{4}\pazocal{C}_{w}^{(k)}\right)_{[3]} \in\R^{NT\times P}$ and $\tilde{\mA}:=\left(\tA\right)_{[3]}\in\R^{NT\times J}$,
      \[
      \pazocal{C}_{h}^{(k+1)}
      \leftarrow\arg\min_{\mC} \tfrac12\left\|\tilde{\mA}-\tilde{\mV}_k\mC^{\top}\right\|_{F}^{2} =\left(\left(\tilde{\mV}_k^{\top}\tilde{\mV}_k\right)^{-1} \tilde{\mV}_k^{\top}\tilde{\mA}\right)^\top.
      \]
      
    \State \textbf{Update \(\pazocal{C}_{w}\):} Form $\bar{\mV}_k := \left(\pazocal{V}\times_{3}\pazocal{C}_{h}^{(k+1)}\right)_{[4]}\in\R^{NTJ\times Q}$ and $\bar{\mathbf{a}} := \vop(\tA)\in\R^{NTJ}$
      \[
      \pazocal{C}_{w}^{(k+1)} \leftarrow \arg\min_{\vc}\tfrac12 \left\|\bar{\va}-\bar{\mV}_k \vc\right\|_{2}^{2} = \left(\bar{\mV}_k^{\top}\bar{\mV}_k\right)^{-1} \bar{\mV}_k^{\top}\bar{\va}.
      \]
    
    \State Compute \(\delta_{h} =\frac{\|\pazocal{C}_{h}^{(k+1)}-\pazocal{C}_{h}^{(k)}\|_{\infty}}{\|\pazocal{C}_{h}^{(k+1)}\|_{\infty}}\), 
    \(\delta_{w}=\frac{\|\pazocal{C}_{w}^{(k+1)}-\pazocal{C}_{w}^{(k)}\|_{\infty}}{\|\pazocal{C}_{w}^{(k+1)}\|_{\infty}}\).
    \If{\(\max(\delta_{h},\delta_{w})<\varepsilon\)} 
    \State break
    \EndIf
\EndWhile

\State 
\(\widehat{\pazocal{B}}_{h}\gets\pazocal{C}_{h}^{(k+1)} \mU_{h}^{\dagger},\quad
 \widehat{\pazocal{B}}_{w}\gets\bigl(\mU_{w}^{\dagger}\bigr)^{ \top}\pazocal{C}_{w}^{(k+1)}.\)
\end{algorithmic}
\end{algorithm}

% \begin{itemize}
%     \item Tensor-Regression Formulation
%     \item Alternating Least Squares (ALS) for $\mathcal{B}_h$ and $\mathcal{B}_w$
% \end{itemize}

\subsection{Residuals Model}\label{section:4.2}
Joint angles for each video frame are estimated based on the derived parameters, and the residuals are  computed as
\[
\vr_t = \va_t - \widehat{\pazocal{B}}_h \mX_t \widehat{\pazocal{B}}_w.
\]
In the classical framework of Gaussian Process Regression (GPR), the noisy model $\vr_t = \mathbf{f}(t) + \epsilon_t$ is employed, wherein the explanatory variable $t\in\mathcal{T}$ is utilized to predict the dependent variable $\vr_{t}\in\R^J$. The latent function \(\mathbf{f}\), which typically remains undetermined, encapsulates trends and dependencies, while $\epsilon_t\sim\pazocal{N}(0,\sigma^2 \mI_{J})$ represents a white noise process. Given a set of residuals \(\mR=\{t_i, \vr_{t_i}\}\), we aimed to determine the conditional probability distribution \(p(\vr_{t_\star}| t_\star,\mR)\) for the following frame. We assume that $t_i\in\mathcal{T}$ is known without noise.

The residuals matrix $\mR =\left[\vr_{t_1}^{\top}, \vr_{t_2}^{\top},\dots, \vr_{t_T}^{\top}\right] ^\top\in \R^{T\times J}$ is defined in the context where $T$ denotes the cycle length. Note that $\mathbf{f}(\cdot)$ is a multivariate Gaussian process on $\mathcal{T}$, characterized by a vector-valued mean function \(m:\mathcal{T}\to\R^{J}\), a covariance kernel \(k:\mathcal{T}\times\mathcal{T}\to\R\), and a positive semi-definite parameter matrix \(\Sigma\in\psd^{J}_+\). Assuming GPR, $\vr_t=\mathbf{f}(t)$ implies $\mathbf{f} \sim \pazocal{MGP}(m, k', \Sigma)$, with $k'(t_i,t_j) = k(t_i,t_j) + \delta_{ij}\sigma^2$, where $\delta_{ij}=1$ holds if $i=j$, and zero otherwise. Let $\mK'$ be the $T\times T$ covariance matrix with \([\mK']_{ij} = k'(t_i,t_j)\). Then, by definition, the collection of functions follows a matrix-variate Gaussian distribution, i.e., $\left[\mathbf{f}(t_1)^\top,\mathbf{f}(t_2)^\top,\dots,\mathbf{f}(t_T)^\top\right]^\top \sim \pazocal{MN}_{T\times J}(\mM,\mK',\Sigma)$.

\paragraph{One-step Ahead Prediction} To predict a new variable $\mathbf{f}_\star$ at the specified testing location $t_\star$, the joint distribution encompassing the observation $\mR=[\vr_1^\top, \vr_2^\top,\dots,\vr_T^\top]$ and the predictive function $\mathbf{f}_\star$ is defined as,
\[
\begin{bmatrix} 
\mR \\[5pt] \mathbf{f}_\star \end{bmatrix} \sim 
\pazocal{MN}_{(T+1)\times J} \left(
\begin{bmatrix}
    \mM \\ \vm_\star^\top
\end{bmatrix}, 
\begin{bmatrix} 
\mK' & \vk'_\star \\[5pt] 
{\vk'_\star}^\top & k'_{\star\star}
\end{bmatrix}, \Sigma \right),
\]
where \(\vm_\star=m(t_\star)\in\R^{J}\), \(k'_{\star\star} = k'(t_\star,t_\star)\in\R\), and $\vk'_\star = \left[k'(t_1,t_\star),k'(t_2,t_\star), \dots, k'(t_T,t_\star)\right]^\top \in \R^T$. Utilizing the conditional distribution of a multivariate Gaussian distribution~\cite{chen2020multivariate}, the predictive distribution is $\mathbf{f}_\star|t_\star,\mR\sim \pazocal{MN}(\widehat{\vm}, \widehat{\Omega}, \Sigma)$, wherein $\widehat{\vm} = \vm_\star +  {\vk'_\star}^\top {\mK'}^{-1} (\mR - \mM)$ and $\widehat{\Omega}= k'_{\star\star} - {\vk'_\star}^\top {\mK'}^{-1} \vk'_\star$ suggest $\mathbf{f}_\star|t_\star,\mR \sim \pazocal{N}(\widehat{\vm}, \widehat{\Omega} \Sigma)$. Furthermore, the expectation and covariance are determined by $\E[\mathbf{f}_\star|t_\star,\mR] = \widehat{\vm}$ and $\Var(\mathbf{f}_\star|t_\star,\mR) = \widehat{\Omega} \Sigma$, respectively.

\paragraph{Mean and Covariance Kernel Functions}
Covariance matrices constitute the core component of MVGP inference. The previously mentioned regression model integrates two covariance matrices, specifically the column covariance and the row covariance. It is important to highlight that only the column covariance is dependent on input and is identified as the kernel covariance function. This function establishes a covariance over a potentially infinite set $\mathcal{T}$ and determines the covariance element between any pair of arbitrary sample locations \cite{Roberts2011}. This is due to its encapsulation of the assumptions regarding the function to be inferred, thereby delineating the proximity and similarity among data points.

Analogous to conventional GPR, the choice of kernels plays a crucial role in determining the performance of multivariate GPR. The squared exponential (SE) kernel is predominantly utilized due to its straightforward formulation and a multitude of beneficial properties, such as smoothness and the capability to integrate with other functions. The function is articulated as follows
\[
k_{SE}(\vx,\vx') = \sigma_s^2 \exp\left(-\frac{\Vert\vx - \vx'\Vert^2}{2 \ell^2}\right),
\]
wherein $\sigma_s$ denotes the signal variance, synonymous with the average deviation of the function from its mean, which is described as the output-scale amplitude. In contrast, $\ell$ relates to the input scale, encompassing dimensions such as length or time.

The prior mean encapsulates our a-priori expectation of the residual trajectory in the absence of attack. Because each operation cycle is expected to follow the same nominal path, an intuitive and statistically efficient choice is the empirical mean computed across the \(N\) normal-operation replications collected offline. We set
\[
m(t)=\frac{1}{N}\sum_{i=1}^{N}\vr_{t}^{i},\qquad t\in\{1,\dots,T\}.
\]
Centreing the MVGP on \( m(t)\) removes repeatable, cycle-level trends, allowing the covariance kernel to focus on stochastic deviations that are informative for detection.

\paragraph{MVGP Parameter Estimation} Let the covariance kernel be denoted by $k(t_i,t_j;\Theta)$, with $\Theta=\{\sigma_s,\ell\}$ characterizing its parameters. Consequently, we obtain $\mK'=\mK_{\Theta}+\sigma^2 \mathbf{I}_{T}$. The objective is to estimate the parameters of MVGP, $\Theta$, $\sigma$, and $\Sigma$. The maximum likelihood estimation (MLE) is an appropriate approach for estimating these parameters. Assume that we have collected $N$ replications of the identical operations, namely $\tX_1, \dots, \tX_N$ and $\mA_1, \dots,\mA_N$. For every replication, we compute $\mR_i = \mA_i - \tX_i\times_2 \widehat{\pazocal{B}}_h \overline{\times}_3 \widehat{\pazocal{B}}_w$ and $i=1,\dots,N$. Since these replications are $i.i.d$, the joint likelihood function of the matrix-variate normal distribution is defined as
\[
p\left(\left\{\mR_i\right\}_{i=1}^{N} \bigg|\mM, \mK',\Sigma\right)= \prod_{i=1}^{N} \frac{\exp\left(-\frac{1}{2}\Tr\left(\Sigma^{-1} \left(\mR_i - \mM\right)^\top (\mK')^{-1} (\mR_i - \mM)\right)\right)}{(2\pi)^{\frac{TJ}{2}} |\Sigma|^{\frac{T}{2}} |\mK'|^{\frac{J}{2}}}.
\]
Taking the log-likelihood
\begin{align*}
    \pazocal{L}\left(\mK',\Sigma,\left\{\mR_i\right\}_{i=1}^{N} \right) &= \log p\left(\left\{\mR_i\right\}_{i=1}^{N} \bigg|\mM, \mK',\Sigma\right) \\
    &= - \frac{NTJ}{2}\log 2\pi - \frac{NT}{2} \log |\Sigma| - \frac{NJ}{2} \log |\mK'| -\frac{1}{2} \sum_{i=1}^{N} \Tr\left(\Sigma^{-1} (\mR_i - \mM)^\top (\mK')^{-1} (\mR_i - \mM)\right),
\end{align*}
parameters can be estimated by maximizing the log-likelihood function with respect to $\Theta$, $\sigma$, and $\Sigma$.

\subsection{Online Detection Algorithm}\label{section:4.3}
Algorithm~\ref{alg:online-detector} provides the pseudocode for online cyberattack detection. In an online detection framework, the objective is to contrast each new observed residuals with the predictions of the MVGP model under normal operating conditions. For each new frame or temporal increment $t_\star$, the residuals are computed as
\[
\vr_{t_\star} = \va_{t_\star} - \widehat{\pazocal{B}}_h \mX_{t_\star} \widehat{\pazocal{B}}_w.
\]
Using the trained MVGP, one can compute the predictive distribution for the residuals at time location $t_\star$. As previously derived, the predictive distribution is represented by $\mathbf{f}_\star|t_\star,\mR \sim \pazocal{N}(\widehat{\vm}, \widehat{\Omega} \Sigma)$.

To determine whether the observed residual is anomalous, a statistical measure can be employed to quantify the deviation between the observed residual $\vr_{t_\star}$ and the predicted distribution of the MVGP. A prevalent choice for such a measure is the Mahalanobis distance that is
\[
g_{t_\star} = \frac{1}{\widehat{\Omega}} \left(\vr_{\star} - \widehat{\vm}\right)^\top  \Sigma^{-1} \left(\vr_{\star} - \widehat{\vm}\right).
\]
It is noteworthy that, assuming $\vr_{t_\star}$ is drawn from the predictive distribution, this distance adheres to a chi-squared distribution statistically, that is,
\[
\frac{1}{\widehat{\Omega}} \left(\mathbf{f}_\star - \widehat{\vm}\right)^\top  \Sigma^{-1} \left(\mathbf{f}_\star - \widehat{\vm}\right) \sim \chi_{(J)}^2.
\]
Consequently, to detect the presence of an attack with a confidence level of $\alpha$, this distance is contrasted against a threshold defined by the specified confidence level, denoted as $g_{t_\star} \lessgtr \Tilde{g}_{\alpha}$.

\begin{algorithm}[h]
\caption{ViSTR-GP: Online Cyberattack Detection Algorithm}
\label{alg:online-detector}
\hspace*{\algorithmicindent}\textbf{Input:} 
Estimated parameters $\widehat{\pazocal{B}}_{h},\widehat{\pazocal{B}}_{w}$; 
MVGP parameters $(m, \mM,\mK',\Sigma, \mR)$;  
decision threshold $\tilde g_{\alpha}$.\\
\hspace*{\algorithmicindent}\textbf{Output:} 
alarm flag $\textsc{Alarm}(t)\in\{\text{True},\text{False}\}$ for each time location $t_\star$.
\vspace{2pt}
\begin{algorithmic}[1]
\For{$t_\star = 1,2,\dots$ }\Comment{streaming time index} 
      \State Acquire raw camera frame $\mathbf{F}_{t_\star}$ and encoder signal $\va_{t_\star}\in\R^{J}$.
      \State Get masked-frame: $\mX_{t_\star} \leftarrow\textsc{SAM-Track}(\mathbf{F}_{t_\star})$.
      \State Compute vision-based state estimation: $\widehat{\va}_{t_\star} \leftarrow \widehat{\pazocal{B}}_{h} \mX_{t_\star} \widehat{\pazocal{B}}_{w}$.
      \State Compute residuals: $\vr_{t_\star} \leftarrow \va_{t_\star}-\widehat{\va}_{t_\star}$.
      \State Compute MVGP predictive distribution parameters
             \[
             \widehat{\vm}_{t_\star} = \vm_\star + {\vk'_\star}^\top {\mK'}^{-1} (\mR - \mM),\qquad
             \widehat{\Omega}_{t_\star} = k'_{\star\star} - {\vk'_\star}^\top {\mK'}^{-1} \vk'_\star.
             \]
      \State Compute test statistics: $g_{t_\star} \leftarrow 
             \dfrac{1}{\widehat{\Omega}_{t_\star}} 
             (\vr_{t_\star}-\widehat{\vm}_{t_\star})^{ \top}\Sigma^{-1}
             (\vr_{t_\star}-\widehat{\vm}_{t_\star})$.
      \State Decision rule:
             \[
             \textsc{Alarm}(t_\star) = 
             \begin{cases}
                \text{True}, & g_{t_\star}>\tilde g_{\alpha},\\
                \text{False},& \text{otherwise}.
             \end{cases}
             \]
\EndFor
\end{algorithmic}
\end{algorithm}

\section{Experimental Setup and Data Collection}\label{section:5}
\begin{figure}[!ht]
    \centering
    \includegraphics[width=1\linewidth]{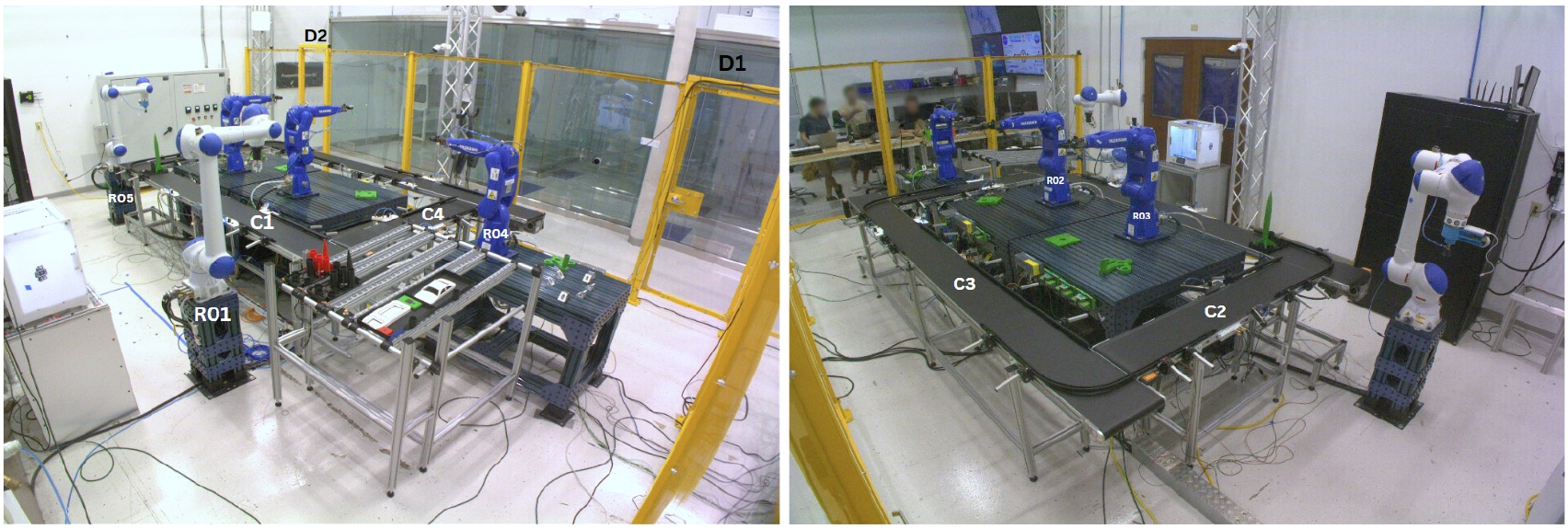}
    \caption{Future Factories Laboratory Testbed Setup.}
    \label{fig:testbed}
\end{figure}

Experiments and data collection were carried out in the Future Factories Laboratory at the University of South Carolina. Figure~\ref{fig:testbed} presents the configuration of the Future Factories Testbed, which integrates a material handling station, a conveyor system, and a network of five Yaskawa industrial robotic arms designed for automated operations. 

The Future Factories Testbed is equipped with five Yaskawa six-axis robotic arms, comprising two HC10s and three GP8s, managed by YRC1000 and YRC1000micro controllers. Engineered for high-speed, high-precision operations, these robots collaborate to perform various assembly tasks. The HC10 units are designated for handling material entry and exit, while the GP8 robots carry out assembly and disassembly operations at dedicated workstations. Each arm is fitted with a uniquely engineered, 3D-printed gripper tailored to its specific function. The GP8 robots have a repeatability precision of 0.01mm. The testbed incorporates a closed-loop system of four interconnected conveyors that encircle all robotic stations. Utilizing C4N Conveyor Belts and Stands, the system enables smooth transfer of products between workstations. Motion control is achieved through two Sinamics GS120 Variable Frequency Drives (VFDs), which are integrated with a PLC. This conveyor setup is fundamental to enabling coordinated interaction among the robotic arms.

Device-to-device and machine communication within the testbed is managed by a Siemens S7-1500 PLC. Configuration, commissioning, and control logic programming are carried out using Siemens’ Totally Integrated Automation (TIA) Portal, which provides a unified environment for engineering and deployment. The PLC executes the control logic essential for orchestrating the assembly process and interfaces with connected systems via multiple communication protocols. Three Distributed I/O modules extend the PLC’s input/output capabilities and can be modularly scaled as needed. Communication with these I/O modules, as well as with the robot controllers and conveyor VFDs, is facilitated through the Profinet protocol.

An Intel RealSense LiDAR L515 camera is used to capture high-resolution video footage of the robot during operation. This depth-sensing camera provides both RGB and 3D spatial data, enabling detailed visual monitoring of the robot's movements. The recorded videos are synchronized with the robot’s joint angle data retrieved from the PLC, allowing for precise correlation between visual behavior and kinematic states. This integration supports tasks such as motion analysis, anomaly detection, and the development of vision-based diagnostics. The camera can record with a resolution of $1920\times1080$ pixels at a rate of 30 frames per second (FPS).

For this paper, robot R04 was taken under consideration to perform the tasks and collect the required data. The use case utilized in the laboratory involves the assembly and disassembly of a rocket prototype composed of four parts. The grippers were specifically designed to grasp components from both the interior and exterior, depending on the requirements of the task. The prototype parts were manufactured with a tolerance of $\pm2 \mathrm{mm}$, making precise robotic motion critical to avoid misalignment or incorrect placement, which are issues that could lead to defects and disruptions along the manufacturing line.

A systematic approach was employed for data collection. Robot R04 was programmed to execute specific tasks while relevant data was gathered. The robot is programmed by manually jogging it using the teach pendant to define key intermediary positions for the end effector. These target locations are recorded, and the motion type between each pair of points is specified, either linear or joint. Linear motion ensures the end effector follows a straight path between positions, while joint motion prioritizes efficiency by coordinating the simultaneous movement of all joints to minimize travel time. Each series of instructions is recorded into a Job with a specific name assigned to it; then these jobs can be executed from the Human-Machine Interface (HMI) manually or executed automatically from a PLC. 

\begin{figure}[!ht]
    \centering
    \includegraphics[width=.9\linewidth]{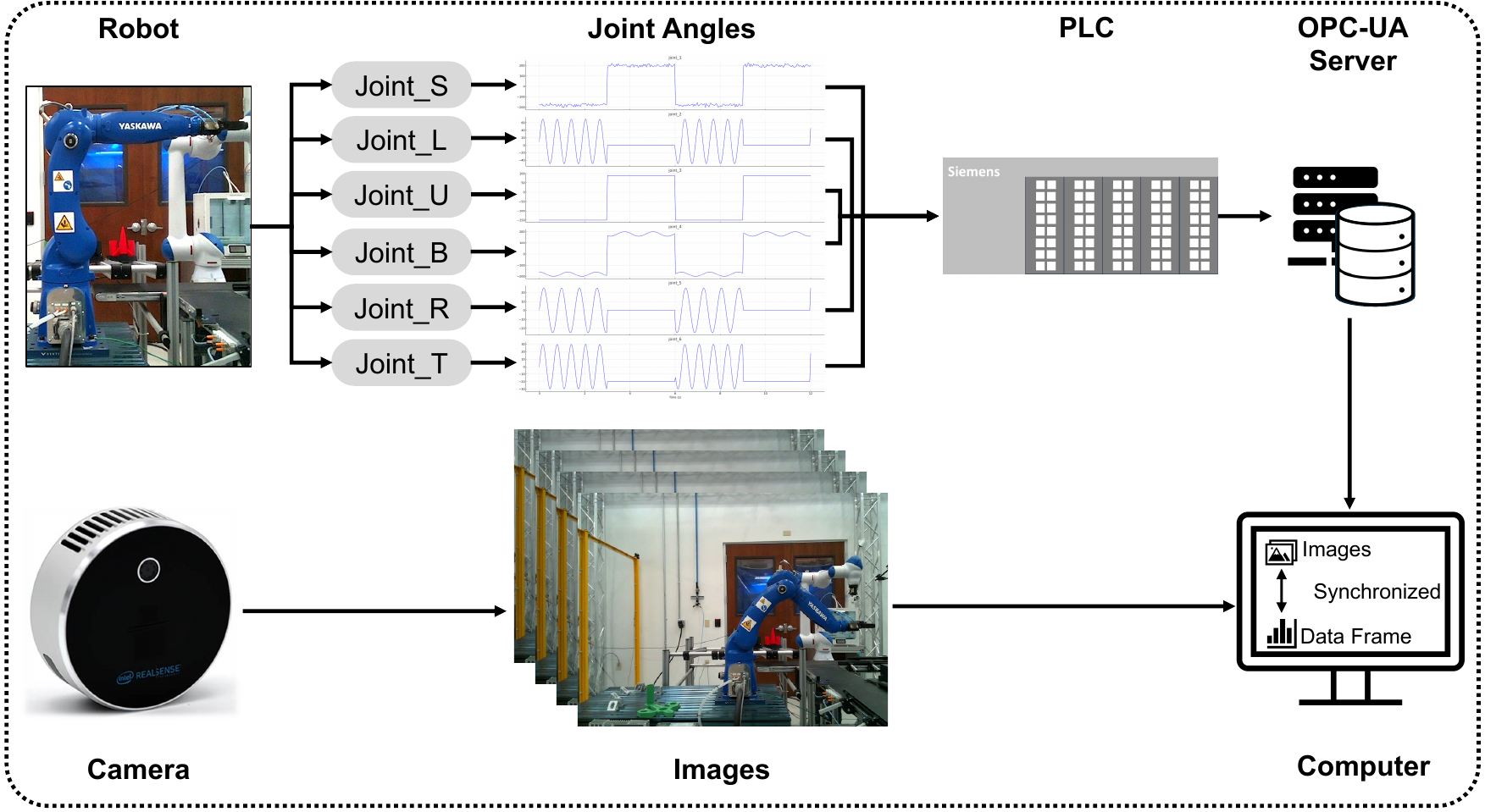}
    \caption{Communication framework used to synchronize robot joint angle data with the image acquisition system.}
    \label{fig:Framework}
\end{figure}

During robot operation, joint angles can be accessed and monitored via ``TIA Portal''. However, direct data logging is not supported through this interface. To enable real-time, structured data acquisition, an OPC-UA server was deployed within the lab’s network. Simultaneously, visual data was captured using a strategically placed Intel RealSense L515 LiDAR camera, which recorded the robot’s movements. The data streams from both the OPC-UA server and the camera were synchronized, ensuring that each video frame could be precisely aligned with the corresponding joint angle data. Figure~\ref{fig:Framework} illustrates the communication architecture used to record and synchronize the robot’s joint angle data and the camera system.

% \begin{figure}
%     \centering
%     \includegraphics[width=1\linewidth]{Framework1.png}
%     \caption{Enter Caption}
%     \label{fig:enter-label}
% \end{figure}

The robot was meticulously programmed to perform tasks that are aligned with the aims outlined in this manuscript. Initially, the robot performed these tasks under normal operating conditions, with no cyberattacks or malicious data injections present. Data was recorded over multiple replications of the task to ensure a sufficient dataset for analysis. The recording system was configured to capture data at a frequency of 30Hz, allowing for high-resolution tracking of the robot’s movements and behavior. Data collection during normal robot operation was followed by creating scenarios that replicate conditions of malicious manipulation. A range of tasks was executed, and corresponding data was collected. These tasks varied from completely different trajectories compared to the baseline job to subtle path alterations that were imperceptible to the naked eye. While identifying significant deviations was straightforward, the primary challenge lay in detecting the minimal, covert changes that could still impact manufacturing quality.

\begin{figure}[!ht]
    \centering
    \includegraphics[width=\linewidth,page=1]{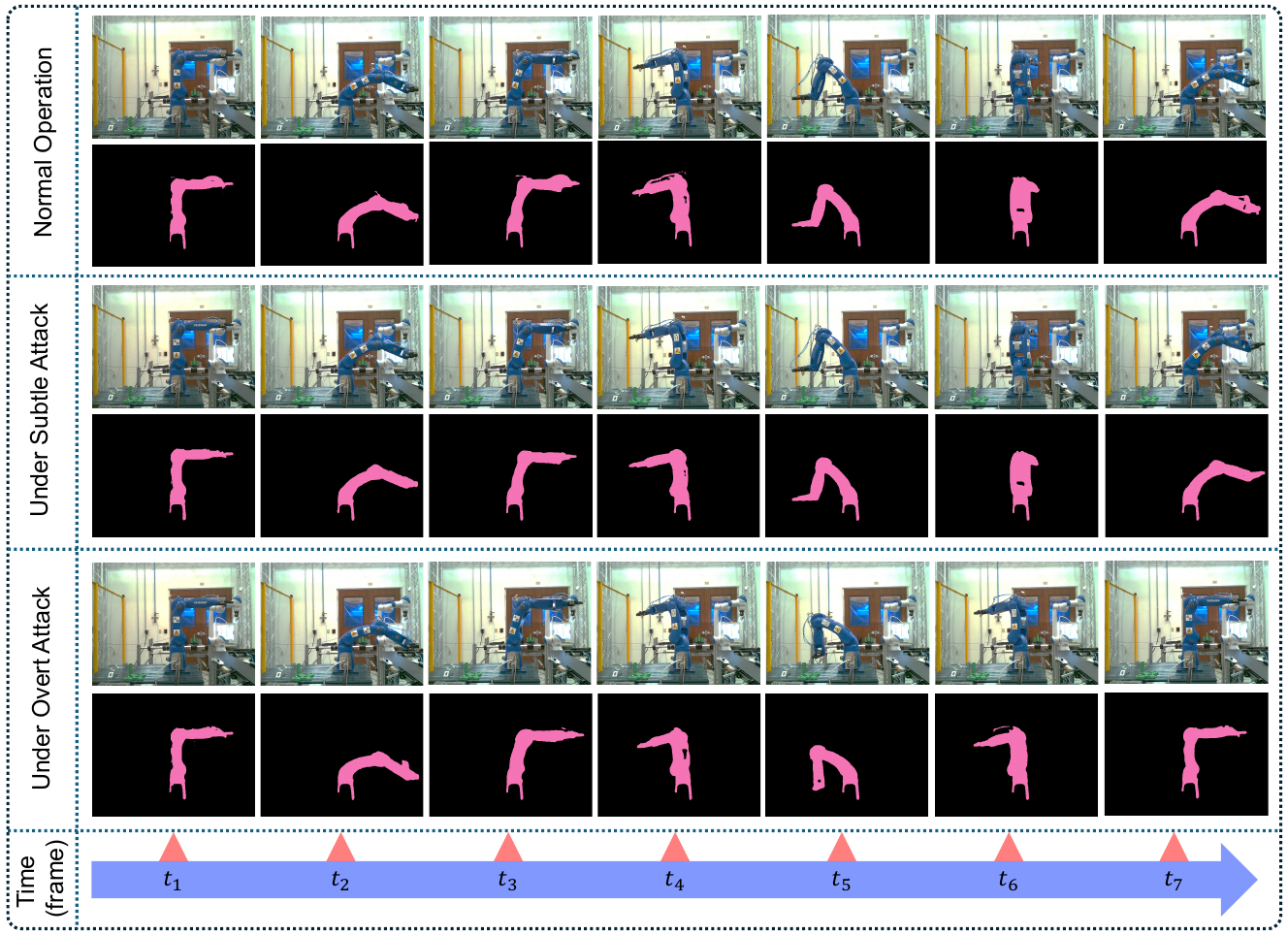}
    \caption{Snapshots from one nominal (top two rows), under subtle attack (middle two rows), and under overt attack (bottom two rows) operation cycle and their corresponding SAM-Track foreground masks.}
    \label{fig:frames}
\end{figure}

The subtly manipulated jobs were designed to mimic the normal task, with some subtle differences and having a deliberate deviation in the robot gripper’s final position. To illustrate the attack scenarios, a task was performed under normal operational conditions, with a subtle attack involving a minor deviation in the end-effector's position, as well as an overt attack wherein the end-effector drifts and follows another trajectory. Figure~\ref{fig:frames} shows representative frame sequences alongside their corresponding SAM-Track masks for nominal and attack cycles. In the initial frame of both the nominal and attack cycles, three user interactions were input into SAM-Track to outline the robot arm. DeAOT extended the mask across all frames and cycles without further human input. The subtle cyberattack is visually undetectable in comparison to a nominal cycle. Conversely, the overt attack becomes apparent subsequent to time $t_4$, as the robot deviates from its standard path and adopts an alternative trajectory.

\begin{figure}[!ht]
    \centering
    \includegraphics[width=\linewidth]{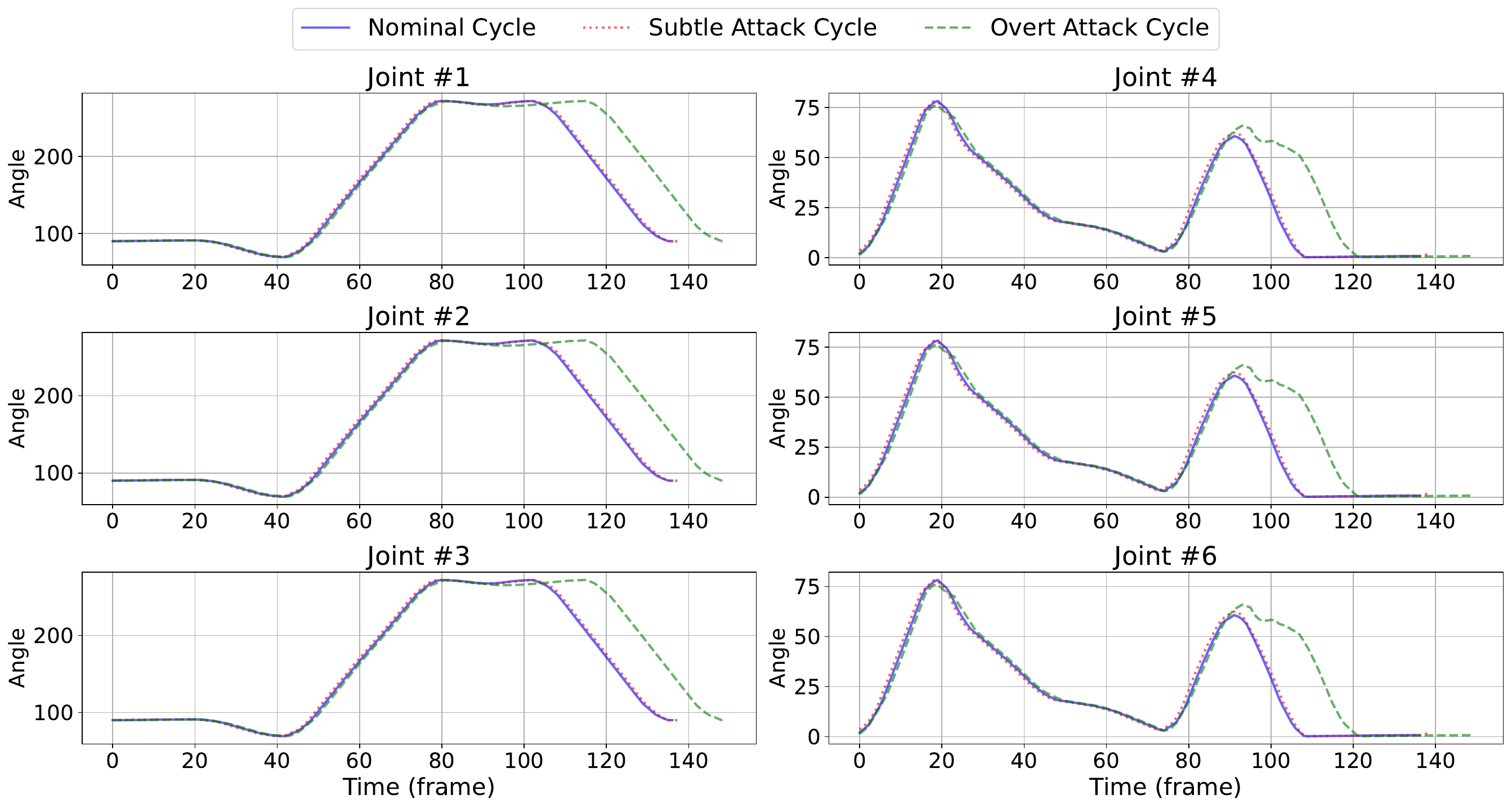}
    \caption{Measured joint angles of nominal (blue), under subtle attack (red), and under overt attack (green) operation cycle. Under replay and covert-type attacks detector observe the measurement of normal operation cycle (blue).}
    \label{fig:angle}
\end{figure}

Figure~\ref{fig:angle} summarizes the encoder-reported trajectories for joint angles corresponding to nominal cycles and cycles subjected to subtle and overt attacks. The diagram illustrates the difficulty in differentiating joint angles in normal operations from those in attack scenarios, especially subtle ones. In the case of an overt attack, the deviation is more evident. Nevertheless, in the context of replay and covert-type attacks, the joint angles observed by the detector correspond to those of a nominal operation cycle (blue lines in Figure~\ref{fig:angle}). This correspondence thereby misleads the detector. Meanwhile, the actual trajectory is subject to manipulation. 

\section{Experiments and Results}\label{section:6}
This section empirically validates the proposed framework on the robotic work-cell data introduced in Section~\ref{section:4}. Section~\ref{section:5.1} details implementation specifics; Section~\ref{section:5.2} introduces the baselines for benchmarking; and Section~\ref{section:5.3} reports the quantitative results and discusses the observed performance and comparison between the proposed framework and the baselines.

\subsection{Implementation Setup}\label{section:5.1}
All experiments were conducted on the robotic work-cell described in Section~\ref{section:3}. Each assembly cycle lasts \(T\approx634\) frames, with $H=480$ and $W=640$, sampled from an overhead camera and synchronized with the robot’s encoder signals. We collected \(N=10\) nominal (attack-free) replications for the purpose of model fitting and its validation, and generated four distinct attack scenarios, each replicated three times. The attack scenarios were designed by introducing deviations in various magnitudes ranging from 5 cm, partly visible, to 0.2 cm, which are hard to visually detect and represent subtle variations. Table~\ref{tab:attacks_scenario} represents the complete list of the utilized manipulated jobs with their respective information.

\begin{table}[!ht]
\centering
\caption{Designed manipulation scenarios. Variation 1 is easily visible and Variation 4 is hard to visually detect.}
\label{tab:attacks_scenario}
% \resizebox{\linewidth}{!}{
\begin{tabular}{l l l l l l}
\toprule
\textbf{Cycles} & & \textbf{Variation 1}& \textbf{Variation 2}& \textbf{Variation 3}& \textbf{Variation 4}\\
\hline
\textbf{Description} & & Downward 5 cm & Downward 1 cm & Downward 0.5 cm & Downward 0.2 cm \\
\bottomrule
\end{tabular}%}
\end{table}

To estimate the TR model outlined in Eq.~\eqref{eq:tr_problem}, each raw frame \(\mathbf{F}_t\) was transformed into a foreground mask image \(\mX_t\in\R^{H\times W}\); these masks were then aggregated to form \(\tX_i\in\R^{T\times H\times W}\) for replication \(i=1,\dots,N\). No additional spatial augmentation was undertaken. Initially, using the \(N\) nominal replications, each \(\tX_i\) was aggregated into $\tY$ and subsequently compressed using a Tucker decomposition. The ranks \((P,Q)\) were selected such that they preserved more than \( 95\%\) of the Frobenius norm of \(\tY\). The reduced representation \(\pazocal{V}\) was then employed to estimate the projection matrices \(\widehat{\pazocal{B}}_h\) and \(\widehat{\pazocal{B}}_w\) via ALS (Algorithm~\ref{alg:ALS}), maintaining a tolerance of \(\varepsilon=10^{-6}\).

Residuals were computed for all nominal replications. The prior mean \(m(t)\) was established as the empirical average across replications at each frame index (Section~\ref{section:4.2}). A squared-exponential kernel \(k(\cdot,\cdot;\Theta)\), equipped with automatic relevance determination and a parameter \(\sigma^2\), was employed to model measurement noise. The kernel hyperparameters \(\Theta=\{\ell,\sigma_s\}\), noise variance \(\sigma^2\), and output covariance \(\Sigma\) were determined by maximizing the log-likelihood function through gradient ascent (see Section~\ref{section:4.2}). During the inference phase, frames were processed in a sequential manner. At each temporal instance \(t_\star\), the residual \(\vr_{t_\star}\) was evaluated using the Mahalanobis statistic \(g_{t_\star}\) as detailed in (\ref{alg:online-detector}). A predetermined threshold \(\tilde g_\alpha\) was adopted on a per-frame basis from the \(\chi^2_J\) distribution at the specified significance level \(\alpha=0.005\).

\subsection{Baselines for Benchmark}\label{section:5.2}
The proposed framework couples a TR visual surrogate with an MVGP residual model. In order to assess the individual contributions of each component, the model is evaluated against three baselines that systematically exclude the MVGP component and/or the TR mapping. For each method, the decision threshold $\tilde g_{\alpha}$ is selected to ensure that the empirical false-alarm rate on the nominal set remains $\alpha=0.005$.

\begin{enumerate}[label=(\roman*)]
\item \textbf{Baseline 1: Tensor Regression with \textit{i.i.d.} Residuals Model (TR + IID):} We retain the bilinear state estimator $\widehat{\va}_t=\widehat{\pazocal{B}}_{h}\mX_t\widehat{\pazocal{B}}_{w}$, but model residuals as independent and identically distributed (\emph{i.i.d.}) Gaussian vectors $\vr_t^{i}\stackrel{\text{i.i.d.}}{\sim}\pazocal{N}(\vm_1,\Sigma_1)$, for $t=1,\dots,T$ and $i=1,\dots,N$. With this ablation, we exclude the temporal features and dependencies from the residual model. The joint likelihood is defined as
\[
p \left(\{\vr_t^{i}\}\big| \mu_1,\Sigma_1\right)=\prod_{i=1}^{N}\prod_{t=1}^{T}
\frac{\exp \left[-\tfrac{1}{2}(\vr_t^{i}-\vm_1)^{ \top}\Sigma_1^{-1}(\vr_t^{i}-\vm_1)\right]}
{(2\pi)^{J/2}|\Sigma_1|^{1/2}} .
\]
Maximizing this log–likelihood yields the closed–form estimators
\[
\widehat{\vm}_1=\frac{1}{NT}\sum_{i=1}^{N}\sum_{t=1}^{T}\vr_t^{i},\qquad
\widehat{\Sigma}_1=\frac{1}{NT}\sum_{i=1}^{N}\sum_{t=1}^{T}
(\vr_t^{i}-\widehat{\mu}_1)(\vr_t^{i}-\widehat{\mu}_1)^{ \top}.
\]
Then, during online detection, for each new frame $t_\star$, we compute residuals $\vr_{t_\star}=\va_{t_\star}-\widehat{\pazocal{B}}_{h}\mX_{t_\star}\widehat{\pazocal{B}}_{w}$, and the test statistic becomes the Mahalanobis distance
\[
g_{t_\star}=(\vr_{t_\star}-\widehat{\vm}_1)^{ \top}\widehat{\Sigma}_1^{-1}(\vr_{t_\star}-\widehat{\vm}_1)
 \lessgtr  \tilde g_{\alpha},
\]
where $\widehat{\vm}_1$ and $\widehat{\Sigma}_1$ are no longer time-dependent.

\item \textbf{Baseline 2: CNN–based State Estimation with MVGP (CNN + MVGP).} 
This baseline replaces the bilinear TR map by a convolutional neural network (CNN) while keeping the residual modeling identical to Section~\ref{section:4.2}. Let $\tilde{\pazocal{F}}_{\phi}$ denote the CNN with parameters $\phi$. The masked frame $\mX_t\in\R^{H\times W}$ (output of SAM-Track) is passed through $\tilde{\pazocal{F}}_{\phi}$. The CNN comprises three convolutional blocks: (i) {\fontfamily{qcr}\selectfont Conv1} with $1\rightarrow16$ channels, $3\times3$ kernel, padding $1$, followed by {\fontfamily{qcr}\selectfont ReLU} and $2\times2$ max pooling; (ii) {\fontfamily{qcr}\selectfont Conv2} with $16\rightarrow32$ channels, $3\times3$ kernel, padding $1$, {\fontfamily{qcr}\selectfont ReLU}, and $2\times2$ max pooling; and (iii) {\fontfamily{qcr}\selectfont Conv3} with $32\rightarrow64$ channels, $3\times3$ kernel, padding $1$, {\fontfamily{qcr}\selectfont ReLU}, and an adaptive average pooling layer that reduces the spatial map to $1\times1$. The resulting $64$-dimensional vector is flattened and passed to a fully connected regressor layer of size $64\rightarrow J$ with linear output. Training uses the normal-operation dataset only. Frames and joint angles are split $80\%/20\%$ into training/validation sets, mini-batches of size $64$ are optimized with Adam (learning rate $10^{-3}$) under an mean squared error (MSE) loss. Early stopping selects the best model. Given $\widehat{\va}_t^{\text{CNN}}=\tilde{\pazocal{F}}_{\phi}(\mX_t)$, residuals are $\vr_t=\va_t-\widehat{\va}_t^{\text{CNN}}$, and the same MVGP model in Section~\ref{section:4.2} and the $\chi^2$ test in Algorithm~\ref{alg:online-detector} remained unchanged.

\item \textbf{Baseline 3: CNN-based State Estimation with \textit{i.i.d.} Residuals (CNN + IID).}  
This baseline removes both the TR map and the MVGP components. We reuse the CNN described in Baseline~(ii) to obtain $\widehat{\va}_t^{\text{CNN}}=\tilde{\pazocal{F}}_{\phi}(\mX_t)$ and then assume $\vr_t^{i}\stackrel{\text{i.i.d.}}{\sim}\pazocal{N}(\vm_1,\Sigma_1)$, estimating $(\vm_1,\Sigma_1)$ exactly as in Baseline~(i). Detection again relies on $g_{t_\star}$ with $\widehat{\vm}_1,\widehat{\Sigma}_1$.
\end{enumerate}

\subsection{Experimental Results}\label{section:5.3}

\begin{figure}[!t]
    \centering
    \includegraphics[width=\linewidth]{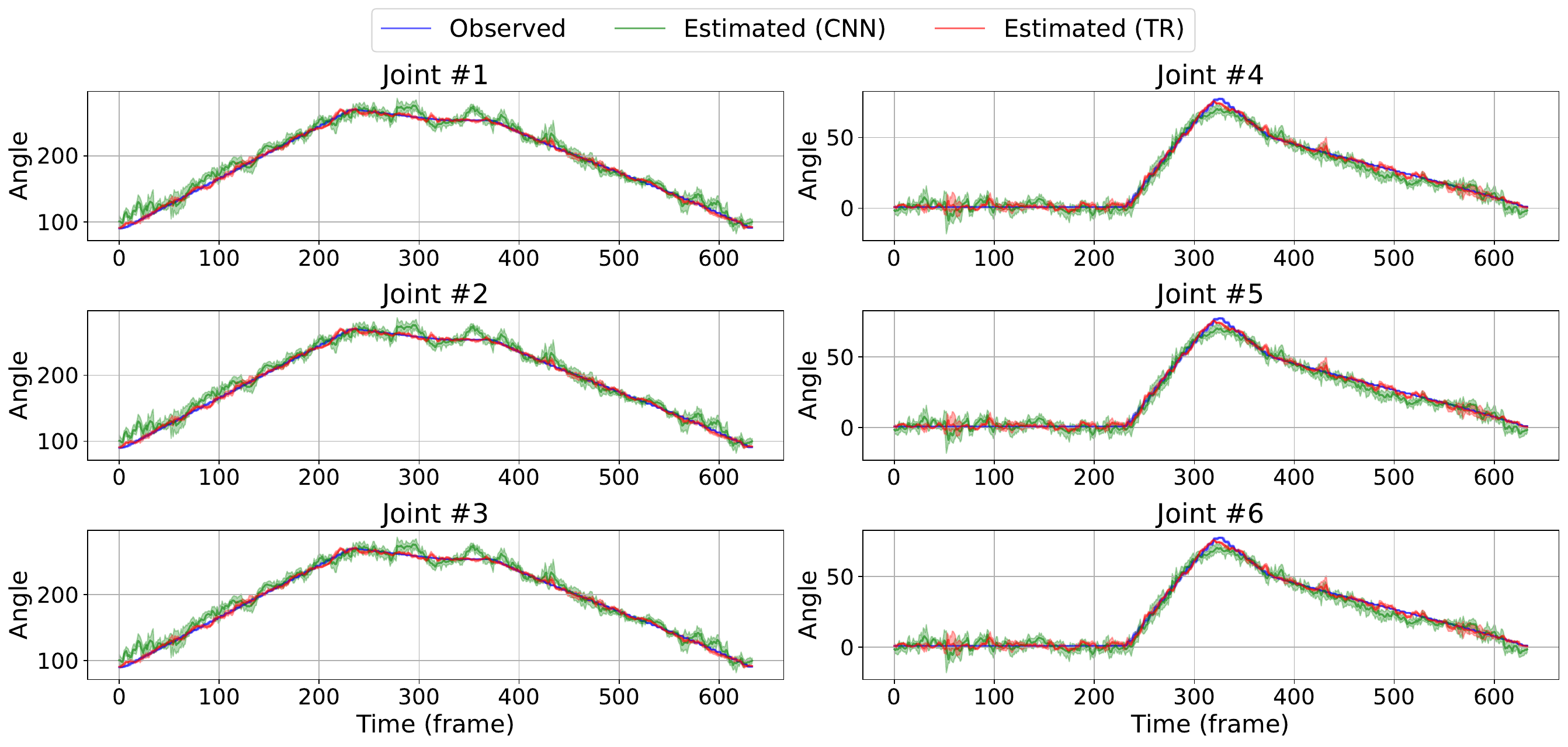}
    \caption{Frame‑wise state–estimation on the full nominal data set for each of the six robot joints. Measured joint angles (blue), estimated joint angles by TR (red) and CNN (green). Solid line is the mean and the shaded area is 2.5\%-97.5\% quantile over replications.}
    \label{fig:angle_comp}
\end{figure}

The results presented in Table~\ref{tab:estimate} and Figure~\ref{fig:angle_comp} indicate that the TR surrogate provides significantly more precise state estimations than the CNN throughout the complete motion range. When averaged across the entire nominal set, TR achieves a root mean squared error (RMSE) of $2.79^{\circ}$ and a mean absolute error (MAE) of $2.09^{\circ}$. In contrast, the CNN incurs errors of $6.19^\circ$ and $4.99^\circ$, respectively. This corresponds to a reduction of approximately 55\% in RMSE and 58\% in MAE. Upon examination of the joint-wise results, two persistent patterns emerge:

\begin{table}[!t]
\centering
\caption{Frame‑wise state–estimation accuracy on the full nominal data set for each of the six robot joints and averaged over all joints. TR is consistently more accurate than CNN, achieving reductions of 55\% in RMSE and 58\% in MAE on average across joints.}
\label{tab:estimate}
\begin{tblr}{
  cells = {c},
  cell{1}{1} = {r=2}{},
  cell{1}{2} = {c=2}{},
  cell{1}{4} = {c=2}{},
  hline{1,10} = {-}{0.08em},
  hline{2} = {2-5}{},
  hline{3,9} = {-}{},
}
Joint \# & RMSE            &        & MAE             &        \\
         & TR              & CNN    & TR              & CNN    \\
1        & \textbf{3.0210} & 9.5119 & \textbf{2.3733} & 7.4902 \\
2        & \textbf{2.0391} & 4.1234 & \textbf{1.5469} & 3.3059 \\
3        & \textbf{1.7602} & 3.5597 & \textbf{1.2892} & 2.9150 \\
4        & \textbf{2.9293} & 6.4251 & \textbf{2.1754} & 5.1984 \\
5        & \textbf{3.4146} & 7.3677 & \textbf{2.5097} & 5.9942 \\
6        & \textbf{3.5566} & 6.1642 & \textbf{2.6163} & 5.0153 \\
Average  & \textbf{2.7868} & 6.1920 & \textbf{2.0851} & 4.9865 
\end{tblr}
\end{table}

\begin{enumerate}[label=(\roman*)]
\item Thorough analysis reveals that the TR model surpasses the CNN model in performance across all joints. This superiority is particularly evident for Joints 1, 4, 5, and 6, which exhibit the most extensive angular trajectories as illustrated in Fig.~\ref{fig:angle_comp}; for these joints, the CNN error is reduced by a factor of 2.1–3.2. Furthermore, even in the case of Joint 3 (characterized by a more limited range as noted in $150^\circ$) the TR model is found to diminish the error by half.

\item The comparison of the MAE by TR to the nominal motion amplitudes in Fig.~\ref{fig:angle_comp} demonstrates that the estimator error consistently remains below $2^\circ-3^\circ$, or less than $4\%$ of the peak angle for all joints. Conversely, the CNN frequently exceeds $10\%$, particularly in the case of Joint 1, where the MAE reaches $7.5^\circ$. This increased bias is directly transmitted into the residuals, consequently inflating the predictive covariance in the subsequent MVGP step.
\end{enumerate}

Integrating a low-rank structure in the TR map aligns well with the robot's planar motion (see Fig.\ref{fig:angle_comp}), leading to more precise and stable state estimates. The residuals resulting from TR estimates display a more informative predictive distribution, as detailed in Table~\ref{tab:mvgp}. This enhanced distribution facilitates the proposed detector's capability to function with narrower uncertainty margins and reduced detection latency.

Let \(\vr_{t}^{b}\in\R^{J}\) denote the residuals generated by a designated state estimator (TR or CNN) within a probabilistic residual framework (MVGP or IID) where $$b\in\{\mathrm{ViSTR-GP}, \mathrm{TR+IID}, \mathrm{CNN+MVGP}, \mathrm{CNN+IID}\}.$$ Each probabilistic residual model yields the predictive mean $\widehat{\vm}_{t}^{b}\in\R^{J}$ and covariance $\widehat{\Sigma}_{t}^{b}\in\psd_{+}^{J}$. MVGP models provide estimates of \(\widehat{\Sigma}_{t}\) for all $t$, thereby capturing the temporal covariance among residuals; in contrast, IID baselines simplify to a time-independent covariance estimate. To evaluate the quality of the residual model performance and their informativeness, the following two metrics are calculated:
\begin{enumerate}[label=(\roman*)]
    \item The sharpness, or log-volume, of a predictive $\widehat{\Sigma}_{t}^{b}$ represents the geometric mean variance. Owing to the fact that $\left| \ \widehat{\Sigma}_{t}\right|$ is proportional to the squared volume of the $J-$dimensional credible ellipsoid, a more negative value signifies a narrower and thus more informative predictive distribution. The computation of this metric is as follows:
    \[
    \mathrm{log‑VOL}=\frac{1}{T}\sum_{t=1}^{T}\frac{1}{2} \log \left|\ \widehat{\Sigma}_{t}^{b}\right|,
    \]
    \item The negative log-likelihood serves as a direct criterion for ranking based on likelihood. Consequently, lower values denote a superior joint fit to center and spread simultaneously. This metric is calculated as follows:
    \[
    \mathrm{NLL} = \frac{1}{T}\sum_{t=1}^{T} \left[\frac{1}{2}\log\left|2\pi\widehat{\Sigma}_{t}^{b}\right| + \frac{1}{2} \left(\vr_{t}^{b} - \widehat{\vm}_{t}^{b}\right)^{\top} \left(\widehat{\Sigma}_{t}^{b}\right)^{-1} \left(\vr_{t}^{b} - \widehat{\vm}_{t}^{b}\right)\right],
    \]
\end{enumerate}

Table~\ref{tab:mvgp} summarizes results regarding the goodness-of-fit metrics. The proposed ViSTR-GP achieves the smallest NLL, indicating that, among all methods, its joint Gaussian predictions assign the largest probability density to the observed residuals. Both CNN variants exhibit notably larger NLLs, confirming that direct convolutional regression of masks to joint angles is statistically less efficient than the low-rank tensor map. ViSTR-GP also attains by far the most negative log-VOL. TR+IID and CNN+MVGP yield larger determinants, while CNN+IID is two orders of magnitude more dispersed. Hence, when judged solely by the concentration of its forecasts, the TR–MVGP combination is the most \emph{informative} model. 

\begin{table}[!ht]
\centering
\caption{Evaluation of residual model fit in the nominal data set.}
\label{tab:mvgp}
\begin{tabular}{lcc}
\toprule
Model                  & NLL     & log-VOL \\
\hline
Baseline 1: TR + IID   & 4.7544  & -3.7587 \\
Baseline 2: CNN + MVGP & 6.4972  & -3.9172 \\
Baseline 3: CNN + IID  & 10.1428 & 1.6297  \\
Proposed: ViSTR-GP    & \textbf{4.6692}  & \textbf{-9.8517} \\
\bottomrule
\end{tabular}
\end{table}

We assess detection performance under a replay attack, a particularly subtle data–integrity attack in networked manufacturing systems. The attacker records a measurement sequence $\{\va_t^r\}_{t=1}^{T}$ from a nominal cycle and, after one clean cycle, begins replaying $\va_t^A=\va_{t-\Delta t}$ while the true motion is physically perturbed at the end–effector. We study downward displacements of $\{0.2,\,0.5,\,1,\,5\}$~cm (Table~\ref{tab:attacks_scenario}); each severity is executed for three replications, each consisting of three attacked cycles following the initial nominal cycle. For every method, a fixed decision threshold is selected on the nominal set so that the in–control false–alarm rate is approximately $\alpha=0.005$.

\begin{figure}[!ht]
    \centering
    \begin{subfigure}[t]{0.5\textwidth}
        \centering
        \includegraphics[width=\linewidth]{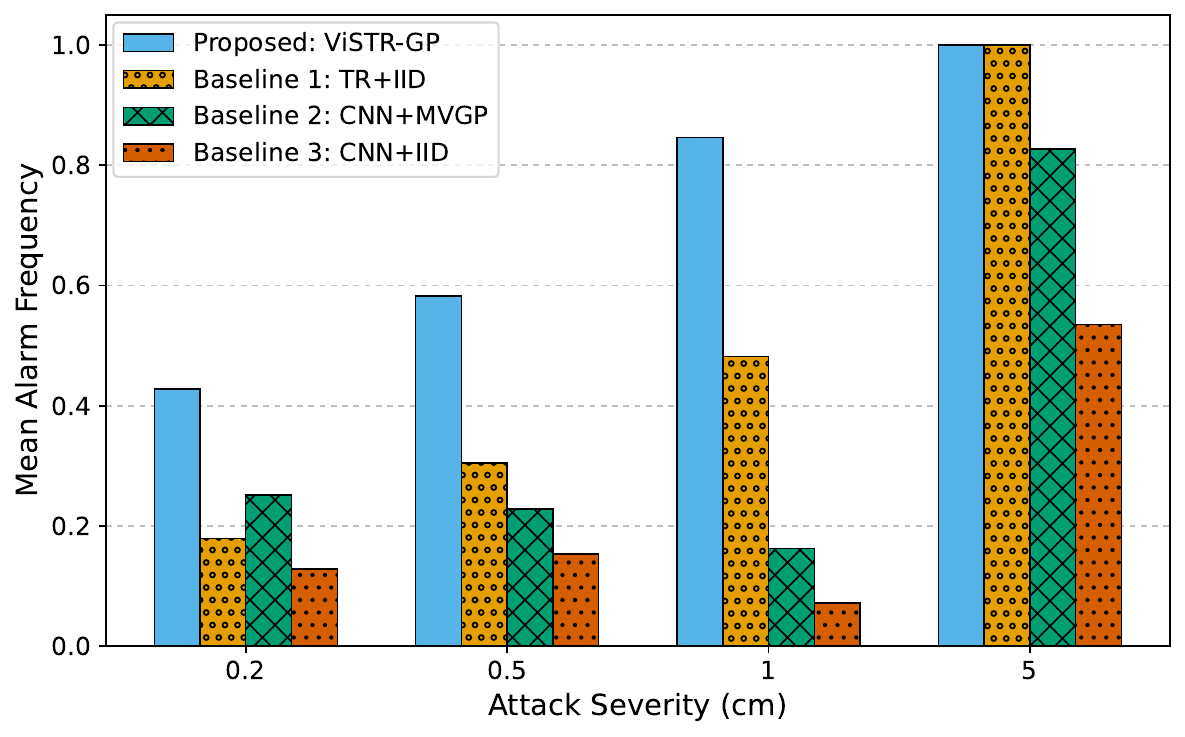}
        \caption{Mean alarm frequency}
        \label{fig:comp1}
    \end{subfigure}%
    \begin{subfigure}[t]{0.5\textwidth}
        \centering
        \includegraphics[width=\linewidth]{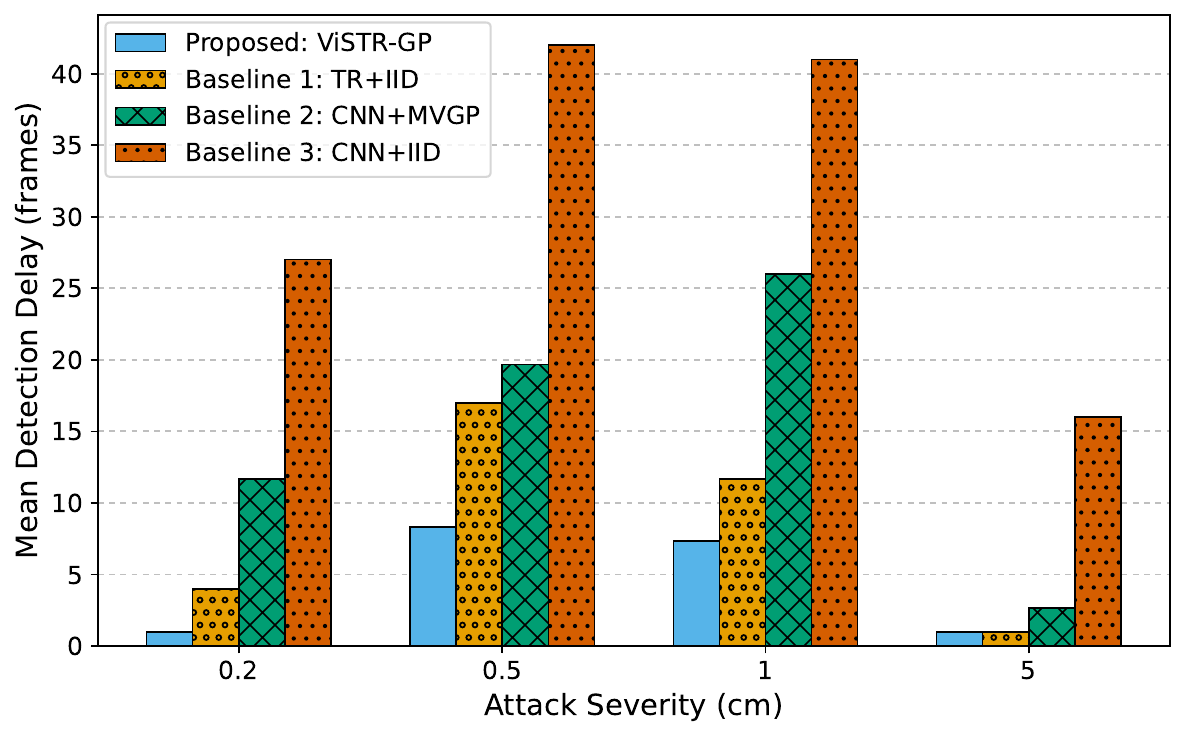}
        \caption{Mean detection delay}
        \label{fig:comp2}
    \end{subfigure}
    \caption{Replay–attack detection performance of the proposed ViSTR-GP against three baselines across four attack severities. (a) reports the mean alarm frequency after the attack onset (higher is better); (b) shows the mean detection delay in frames (lower is better). Bars are averaged over three replications per severity ($\alpha=0.005$ for all methods).}
    \label{fig:comp}
\end{figure}

Figure~\ref{fig:comp} summarizes two complementary metrics aggregated over the attacked cycles. Figure~\ref{fig:comp1} reports the mean alarm frequency after the attack onset (larger is better), computed as the number of frames and measurements flagged within the attacked portion of each cycle and then averaged over the three replications. Figure~\ref{fig:comp2} shows the mean detection delay in frames (smaller is better), defined as the first–passage time from the attack onset to the first alarm, averaged in the same manner. 

Across all severities, the proposed ViSTR-GP yields the highest alarm frequency rate and the smallest delays. The advantage is most pronounced at subtle manipulations (0.2–0.5~cm), where ViSTR-GP raises frequent alarms almost immediately after onset, while CNN+IID exhibits long delay and sparse alarms, and CNN+MVGP remains markedly less sensitive. TR+IID ranks second overall, indicating that exploiting the low-rank bilinear mapping already confers a substantial benefit; however, replacing the IID residual model with MVGP further improves the detection power by leveraging temporal correlation and producing sharper, time-adaptive predictive covariances. As the attack magnitude increases to 5~cm, all methods become easier to trigger (alarm frequencies saturate and delays shrink), yet the proposed detector maintains the best performance, whereas CNN+IID continues to delay by a wide margin. 

\begin{figure}[!ht]
    \centering
    \begin{subfigure}[b]{0.7\textwidth}
        \centering
        \includegraphics[width=\linewidth]{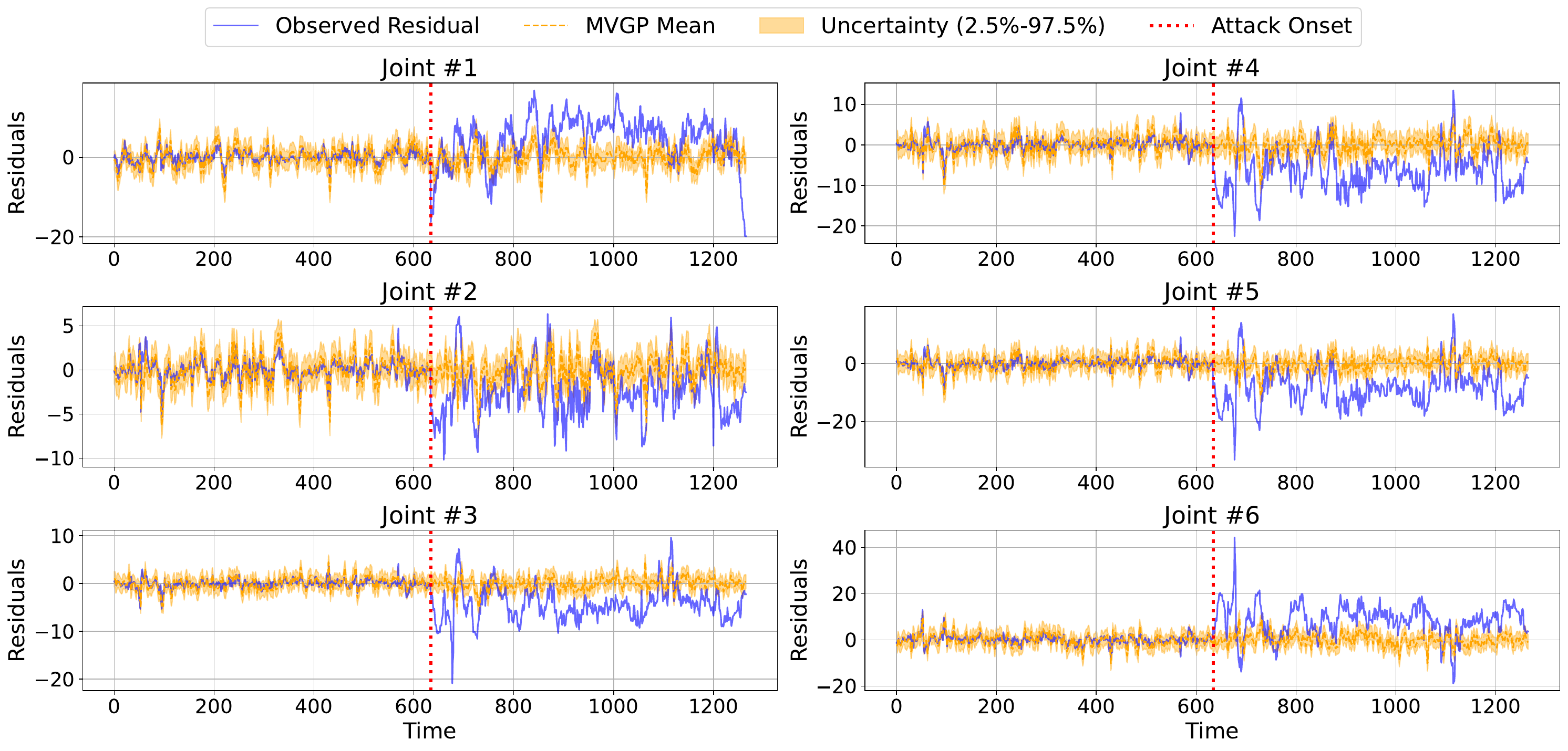}
        \caption{}
    \end{subfigure}%
    \hfill
    \begin{subfigure}[b]{0.3\textwidth}
        \centering
        \includegraphics[width=\linewidth]{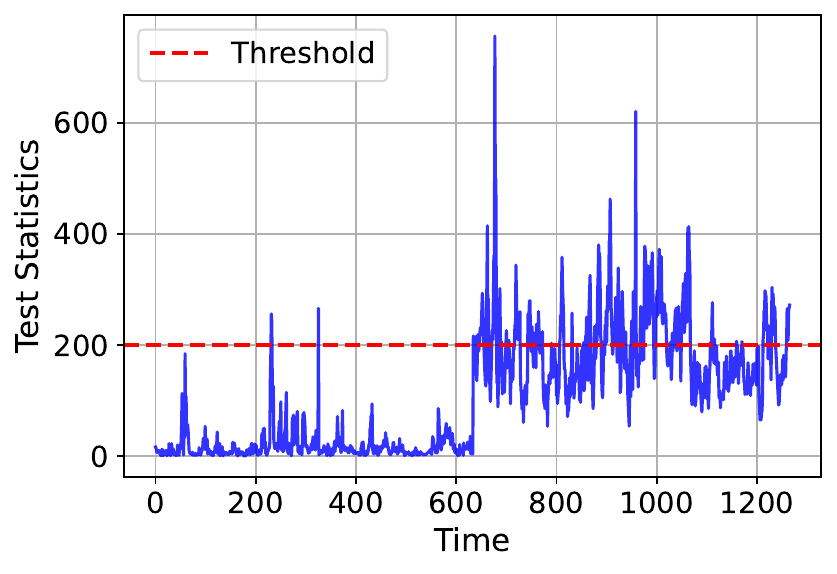}
        \caption{}
    \end{subfigure}
    \caption{ViSTR-GP: Attack 0.2cm. (a) per-joint residuals (blue) with the MVGP predictive mean and uncertainty band (orange), and (b) frame-wise test statistic with the threshold.}
    \label{fig:0.2cm}
\end{figure}

\begin{figure}[!ht]
    \centering
    \begin{subfigure}[b]{0.7\textwidth}
        \centering
        \includegraphics[width=\linewidth]{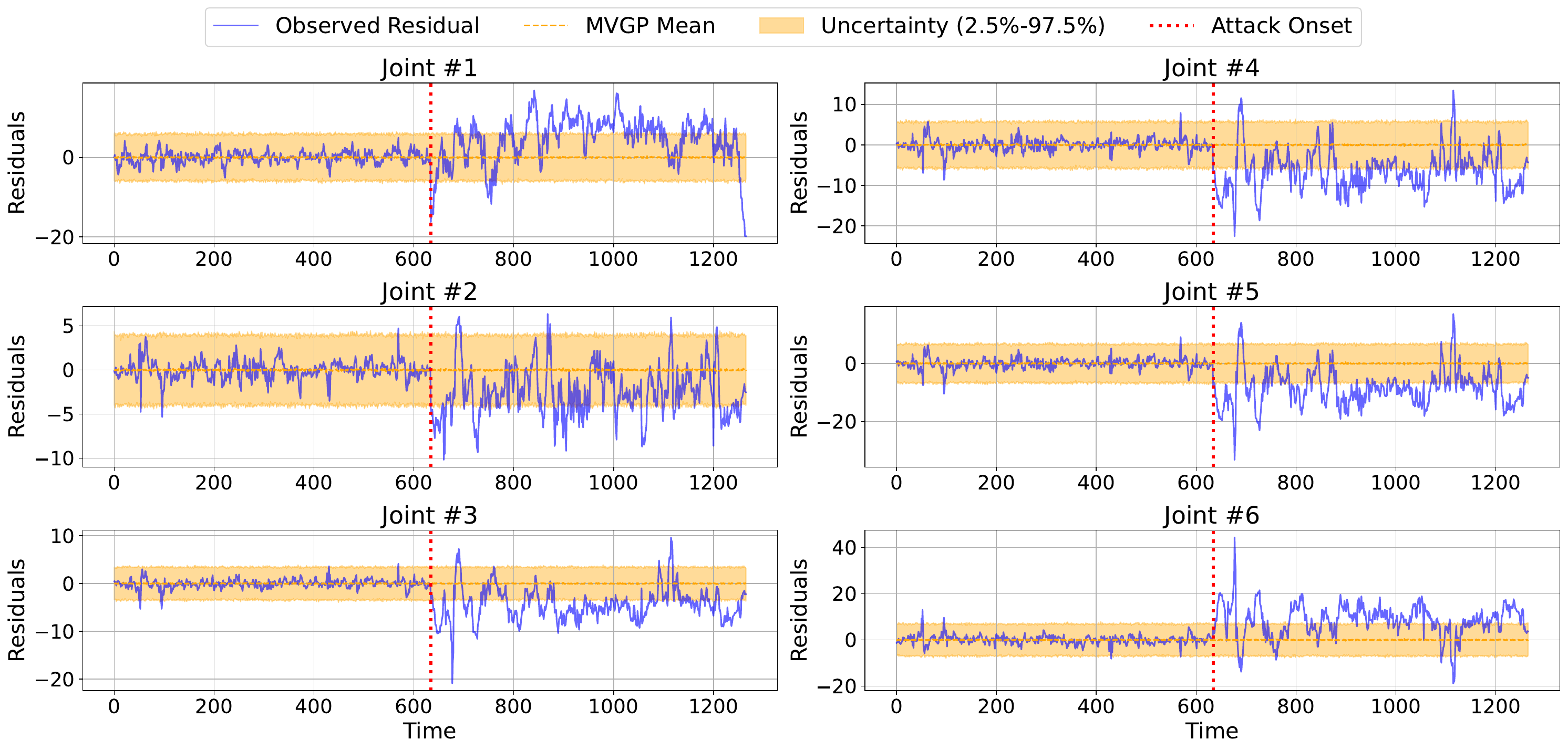}
        \caption{}
    \end{subfigure}%
    \hfill
    \begin{subfigure}[b]{0.3\textwidth}
        \centering
        \includegraphics[width=\linewidth]{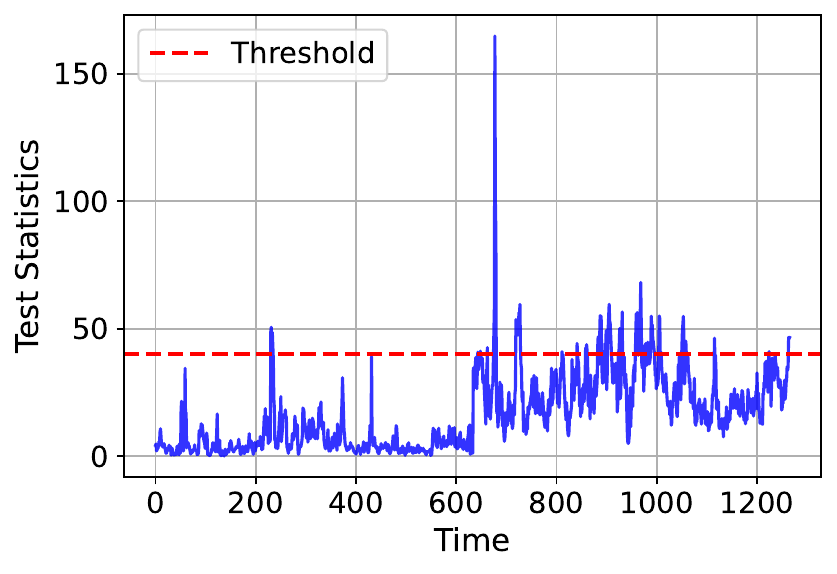}
        \caption{}
    \end{subfigure}
    \caption{TR+IID: Attack 0.2cm. (a) per-joint residuals (blue) with \emph{i.i.d.} mean and uncertainty band (orange), and (b) frame-wise test statistic with the threshold.}
    \label{fig:0.2cm_iid}
\end{figure}

Figures~\ref{fig:0.2cm}-~\ref{fig:0.2cm_iid} illustrate qualitative analyses of residuals and timeline assessments for attack severity 0.2 cm, which substantiate the earlier and more frequent alarms observed with MVGP in comparison to IID. Figure~\ref{fig:0.2cm} shows per-joint residuals (blue) with the MVGP predictive mean and uncertainty band (orange). It also includes the corresponding frame-wise $\chi^2$ test statistic with the threshold. The red dashed line marks the attack onset, separating a normal cycle before and an attack cycle after. Before the attack onset, residuals for all joints are tightly centered within the time-varying MVGP bands. Immediately after onset, systematic drifts emerge from the MVGP predictive distribution, producing sharp spikes in test statistics. Figure~\ref{fig:0.2cm_iid} repeats the analysis when residuals are modeled as \emph{i.i.d.} Gaussian distribution. The empirical mean tracks the pre-attack residuals, but the predictive uncertainty band is time-invariant and noticeably wider than in MVGP. After onset, the residuals shift, yet the inflated, non-adaptive covariance yields slower growth of the $\chi^2$ statistic. Time-dependent MVGP uncertainty reduces false transients in nominal phases while tightening around stable segments, explaining the short delays and frequent alarms. Ignoring temporal structure leads to less informative (wider) bands and fewer alarms, indicating IID's lower sensitivity than MVGP, especially for subtle attacks. Additionally, analogous analyses for alternative attack scenarios are presented in~\ref{section:analysis}.

These trends are consistent with the goodness-of-fit metrics in Table~\ref{tab:mvgp}. The TR surrogate delivers lower estimation error than CNN, reducing residual variance, while MVGP captures the residuals’ temporal dependence, yielding concentrated and more informative predictive distribution. In contrast, CNN-based mappings inflate residual noise through bias, and IID residual models overlook temporal structure, both of which reduce the signal-to-noise ratio available to the detector.

% Qualitative analyses of residuals and timeline assessments for all severity levels, which substantiate the earlier and more frequent alarms observed with ViSTR-GP in comparison to TR+IID, are presented in~\ref{section:analysis}.

\section{Discussion}\label{section:7}
This work introduced an online framework, ViSTR-GP, for cyberattack detection in automated robotic operations. This study demonstrates that an independent, vision-based estimator of robot kinematics can expose subtle data-integrity attacks that evade detectors relying solely on embedded measurements. By mapping per-frame masks to joint angles with a low-rank TR surrogate and modeling nominal residuals with a MVGP, ViSTR-GP achieves rapid, frame-wise cyberattack detection with controlled false-alarm rates. Empirically, the TR surrogate recovers joint angles accurately from a single camera view (RMSE $2.79^\circ$, MAE $2.09^\circ$), substantially outperforming a CNN regressor (RMSE $6.19^\circ$, MAE $4.99^\circ$). The MVGP further yields more informative, time-adaptive uncertainty quantification, enabling earlier detection and more frequent alarms under a replay attack with end-effector deviations as small as $0.2-0.5\mathrm{cm}$.

These findings address challenges outlined in Section~\ref{section:1}. First, to mitigate vulnerabilities in detectors reliant solely on embedded measurements, making them vulnerable to replay and covert-type attacks, we employ a cross-view residual test. This test compares encoder-reported joint angles with camera-based estimates from an external perspective, thereby preventing attacker control over both verification aspects. Secondly, instead of relying on noisy and intrusive side-channels like power monitoring, we employ a fixed overhead camera. By utilizing a one-time SAM-Track segmentation, we achieve clear per-frame masks that effectively minimize background and lighting interference. The detector is exclusively trained on normal cycles, enhancing its scalability when encountering unknown system anomalies. Additionally, no additional instrumentation is necessary. Third, unlike high-fidelity models, which are expensive to maintain and struggle with complex multi-joint nonlinear motions, we utilize a model-free approach. A low-rank bilinear TR map is used to convert masks into joint angles, while a MVGP model captures temporal and cross-joint residual correlations. This allows for clear frame-by-frame test statistics and efficient online processing.

% These findings address three persistent challenges in industrial robot security. First, methods that reason only over controller signals are intrinsically vulnerable to data-integrity attacks; by construction, our detector compares those signals with an external physical channel, breaking the attacker’s control of the evidence stream. Second, physics-based detectors require detailed models and specifications that are costly to maintain across changing recipes and multi-joint nonlinear motions; our approach learns a compact visual surrogate from unlabeled masks and avoids plant-specific dynamics identification while still preserving physical interpretability through joint-angle residuals \cite{Alemzadeh7579758,Hector9473818,Longari10508318,Zhang8835244,Sun9581186}. Third, while side-channel defenses improve robustness, they typically add hardware and integration burden; here, a fixed overhead camera suffices as the independent channel and requires only a one-time, user-verified segmentation to bootstrap tracking.

Two key observations explain the performance differences seen in the experiments. First, the TR surrogate aligns well with the spatial structure of planar robot motion. This alignment yields lower bias and variance than the CNN across all joints. Additionally, this tighter state reconstruction propagates into more concentrated residual distributions and, consequently, higher detection power. Second, explicitly modeling temporal correlation with an MVGP rather than assuming \emph{i.i.d.} residuals produces time-dependent informative predictive distributions that reflect cycle position. This improves both sensitivity immediately after attack onset and robustness to nominal transient phases, whereas \emph{i.i.d.} models trigger late. We expected CNN+MVGP to regain some MVGP benefits despite poorer state estimates, but its alarms were notably sparser than ViSTR-GP at low severities. Post hoc inspection indicates that CNN bias increases the residual baseline, requiring the MVGP to maintain broader predictive spreads, which reduces the advantage of temporal modeling. Joints with larger angular movements increased the performance gap between TR and CNN, reflecting the surrogate’s ability to leverage low-rank structure over wider motion ranges.

The proposed framework has several practical implications for industrial deployment. The detector operates on per-frame statistics with a closed-form decision rule. This allows it to be embedded at the edge, near the camera stream, and monitored centrally. Additionally, this setup requires minimal integration into the control authority. The $\chi^2$ test offers interpretable operating points for quality engineers (e.g., selecting a per-frame significance level to limit false-alarms), while the residuals themselves serve as a diagnostic signal that localizes discrepancies to specific joints. Importantly, the framework complements traditional cyber defenses. It supplies physical corroboration of motion when network data is untrustworthy, strengthening defense-in-depth for standards-driven plants.

Our framework requires a secure camera viewpoint and stable calibration parameters. Any major changes to the camera pose, lens parameters, or severe occlusions necessitate re-segmentation and recalibration. Additionally, ongoing occlusions reduce TR accuracy. Our study evaluated a single robot family and task; generalizing across multiple robots, tools, and highly non-repetitive programs may necessitate multi-view sensing or richer visual features. Finally, although frame-wise testing is simple and effective, we did not employ cumulative change-detection statistics that can further shorten detection delay.

These limitations point to concrete extensions. Fusing camera depth with RGB masks or learning self-supervised geometric embeddings can enhance robustness to lighting and occlusion. In modeling, using kinematic priors in surrogates (e.g., weak joint or link-length constraints) and utilizing state-space MVGPs for scalable inference are promising approaches. Integrating vision with an independent physical channel (e.g., motor current) would enhance security against multi-modal spoofing. Finally, extending the framework to include multiple robots and tasks in operation increases its general applicability.

% \begin{itemize}
%     \item Key findings
%     \item Industrial implications
%     \item Limitations
%     \item Future extensions 
% \end{itemize}

\section{Conclusion}\label{section:8}
This paper introduced an online framework for data-integrity cyberattack detection on automated robotic operations by cross-checking encoder-reported measurements against a camera-based estimate of joint angles. A low-rank TR surrogate maps per-frame masks to joint angles, and a MVGP models nominal residuals to deliver frame-wise $\chi^2$ detection with interpretable thresholds. Across a real robotic testbed with replay attacks, the approach consistently outperformed baselines, recovering joint angles more accurately than a compact CNN and raising earlier, more frequent alarms even under subtle attacks. These results demonstrate a practical path to defense-in-depth. By introducing an independent physical channel that remains outside the controller’s authority, plants can detect data-integrity attacks without adding complex instrumentation. Future work will broaden applicability and scalability by fusing depth with RGB or adding a second view, incorporating lightweight kinematic priors, and adopting state-space GPs for shorter delays, and validating across multiple robots, tools, and tasks. As factories become more connected, this independent visual cross-check offers an immediately deployable step toward certifiable, resilient robotic operations.

\section*{Acknowledgment}
The authors thank Drew Sander for his valuable assistance with data collection.

\section*{CRediT authorship contribution statement}
\textbf{Navid Aftabi:} Conceptualization, Methodology, Software, Formal analysis, Visualization, Investigation, Data Curation, Supervision, Validation, Writing - Original Draft, and Writing - Review \& Editing. \textbf{Jin Ma:} Methodology, Software, and Writing-Original Draft. \textbf{Philip Samaha:} Conceptualization, Data Curation, Software, Writing – Original Draft. \textbf{Ramy Harik and Long Cheng:} Conceptualization, Supervision, Funding acquisition, Project administration, Writing - Review \& Editing. \textbf{Dan Li:} Conceptualization, Methodology, Formal analysis, Investigation, Supervision, Validation, Funding acquisition, Project administration, Writing - Original Draft, and Writing - Review \& Editing.

\section*{Conflict of interest} 
The authors have no conflicts of interest to disclose.

\section*{Funding} 
This work was supported by the U.S. National Science Foundation under Grant 2501479.

\section*{Data Availability Statement}
Data, models, and codes are available upon request.

\bibliographystyle{elsarticle-num} 
\bibliography{ref}

\appendix

\section{Supplementary Analysis}\label{section:analysis}
\begin{figure}[!ht]
    \centering
    \begin{subfigure}[b]{0.7\textwidth}
        \centering
        \includegraphics[width=\linewidth]{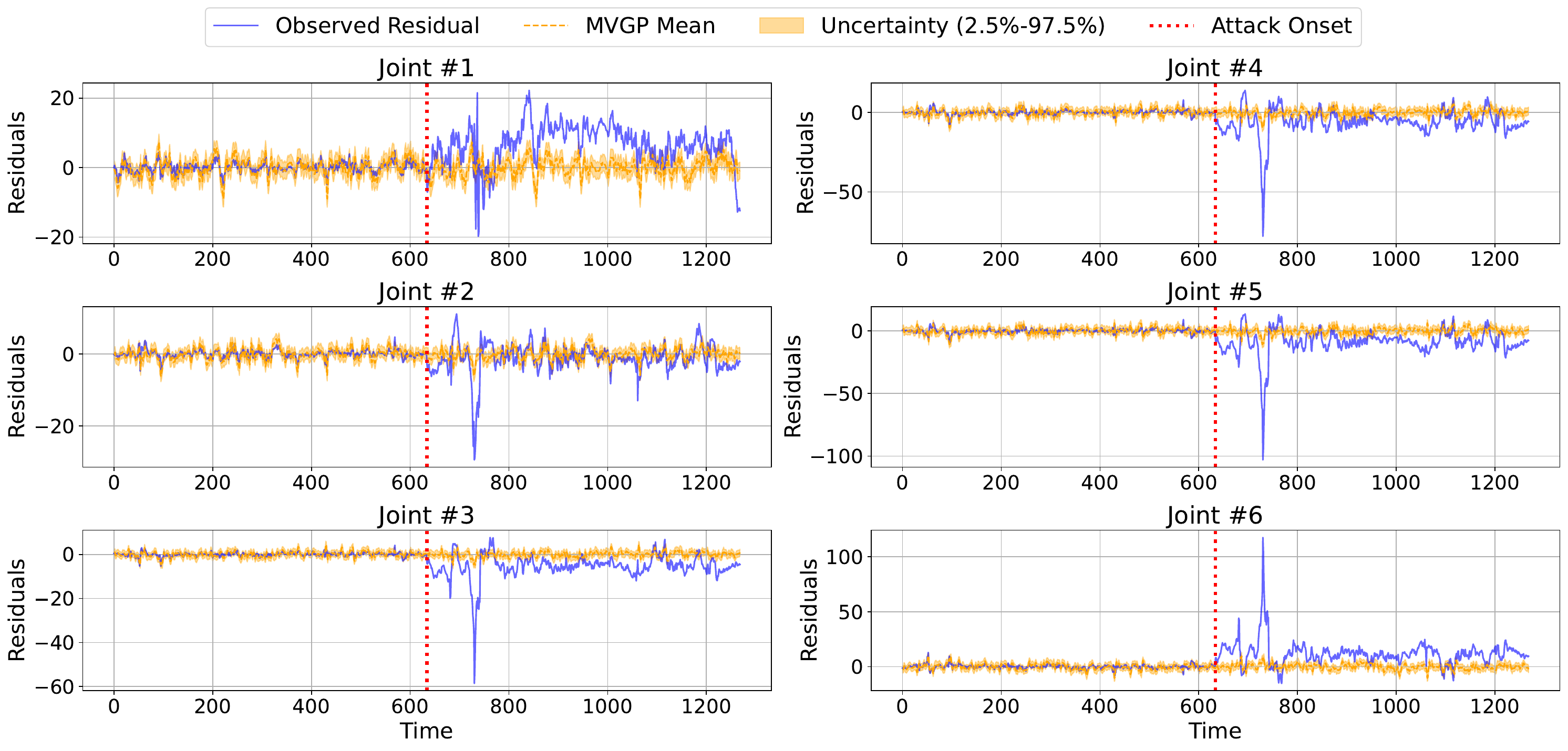}
        \caption{}
    \end{subfigure}%
    \hfill
    \begin{subfigure}[b]{0.3\textwidth}
        \centering
        \includegraphics[width=\linewidth]{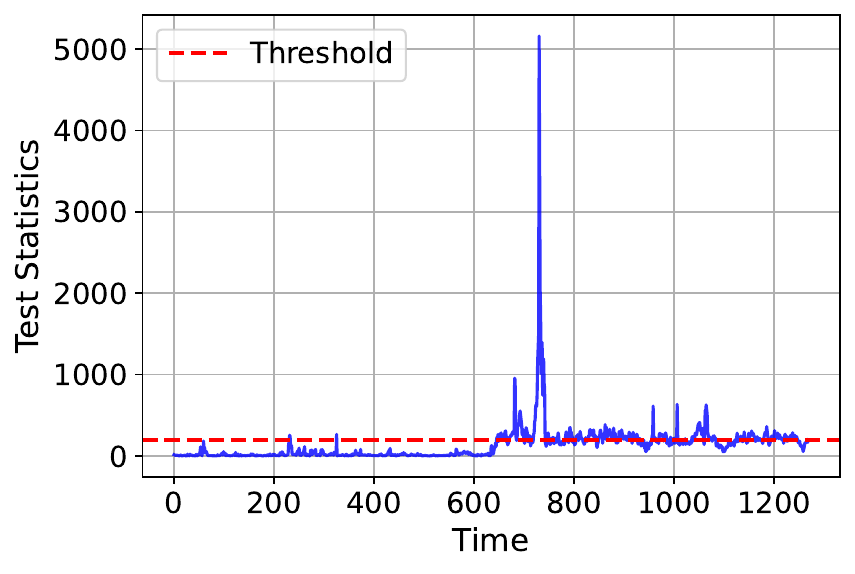}
        \caption{}
    \end{subfigure}
    \caption{ViSTR-GP: Attack 0.5cm. (a) per-joint residuals (blue) with the MVGP predictive mean and uncertainty band (orange), and (b) frame-wise test statistic with the threshold.}
    \label{fig:0.5cm}
\end{figure}

\begin{figure}[!ht]
    \centering
    \begin{subfigure}[b]{0.7\textwidth}
        \centering
        \includegraphics[width=\linewidth]{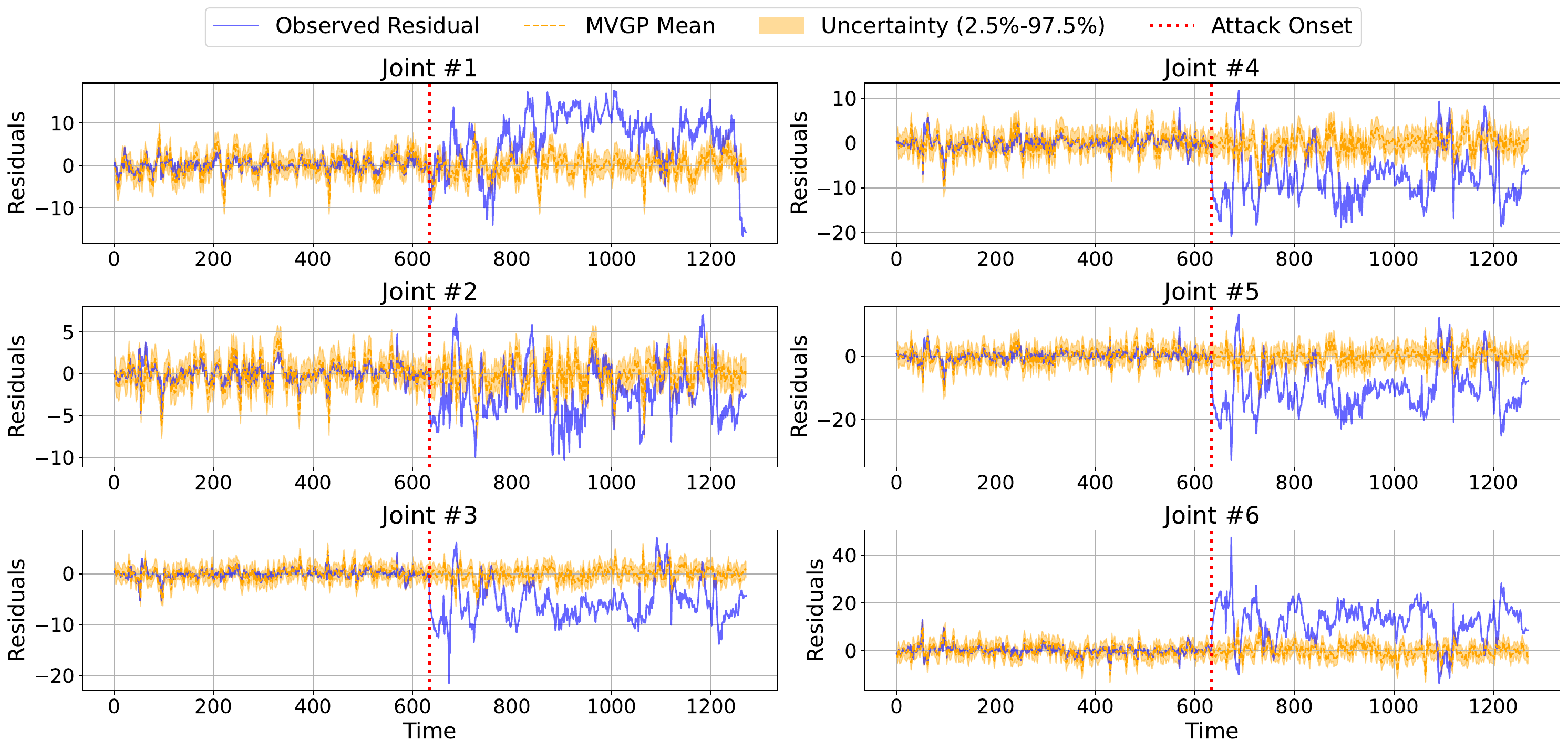}
        \caption{}
    \end{subfigure}%
    \hfill
    \begin{subfigure}[b]{0.3\textwidth}
        \centering
        \includegraphics[width=\linewidth]{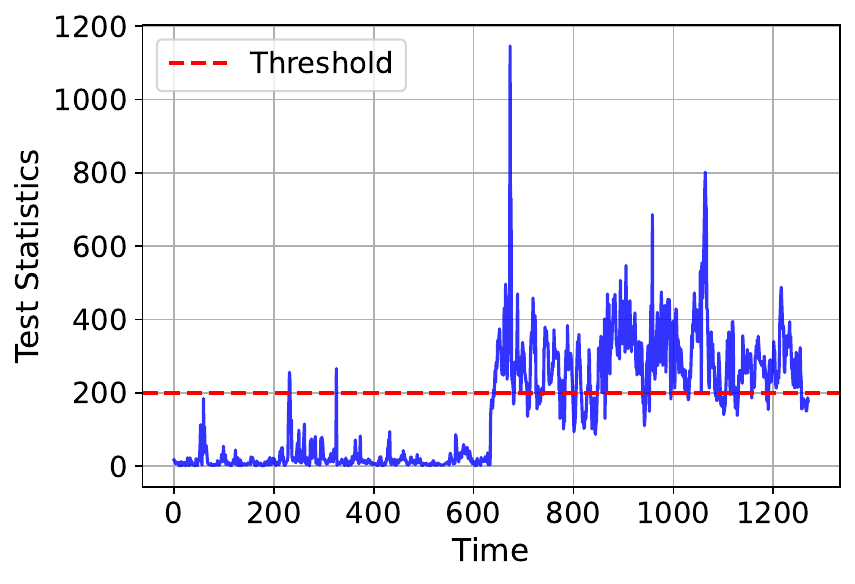}
        \caption{}
    \end{subfigure}
    \caption{ViSTR-GP: Attack 1cm. (a) per-joint residuals (blue) with the MVGP predictive mean and uncertainty band (orange), and (b) frame-wise test statistic with the threshold.}
    \label{fig:1cm}
\end{figure}

\begin{figure}[!ht]
    \centering
    \begin{subfigure}[b]{0.7\textwidth}
        \centering
        \includegraphics[width=\linewidth]{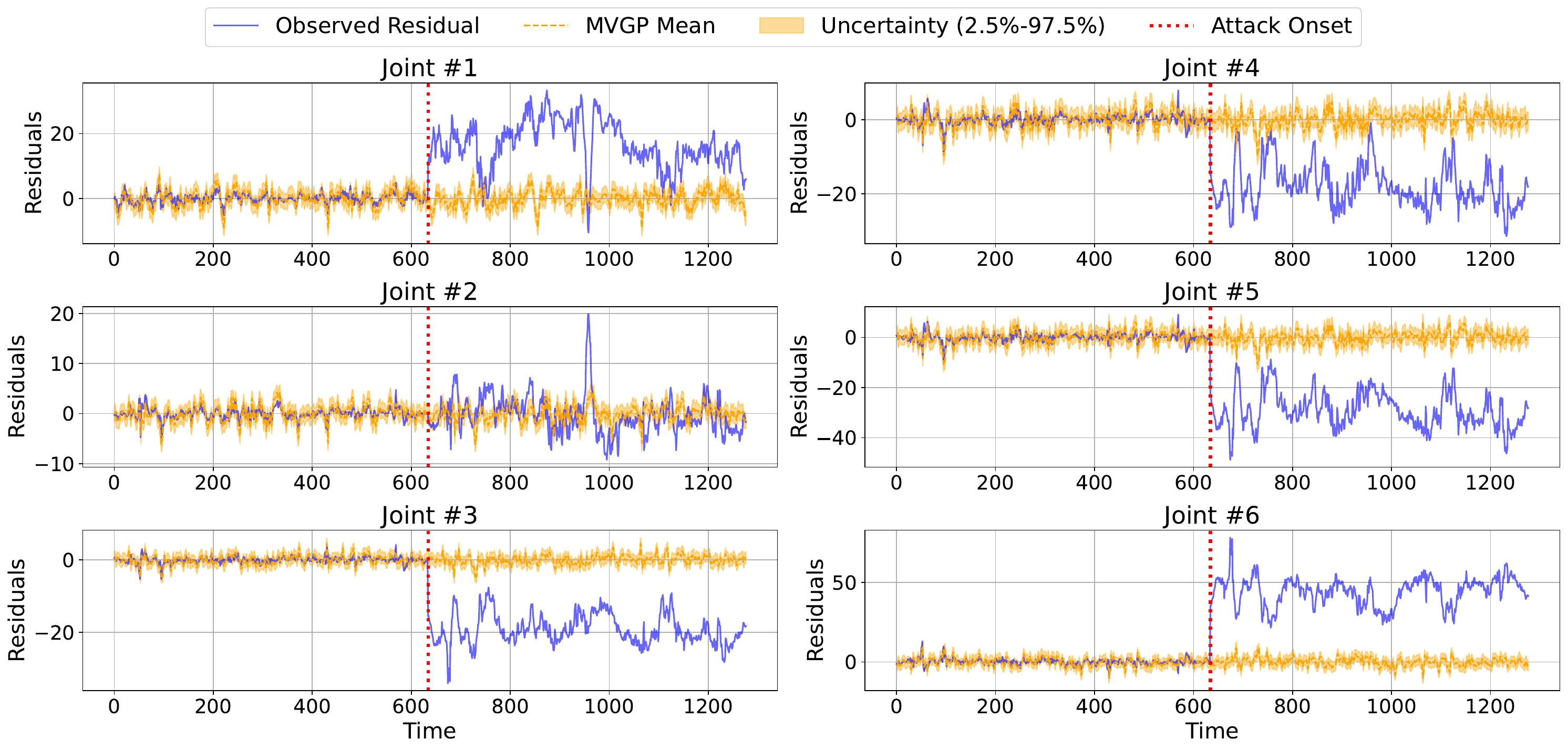}
        \caption{}
    \end{subfigure}%
    \hfill
    \begin{subfigure}[b]{0.3\textwidth}
        \centering
        \includegraphics[width=\linewidth]{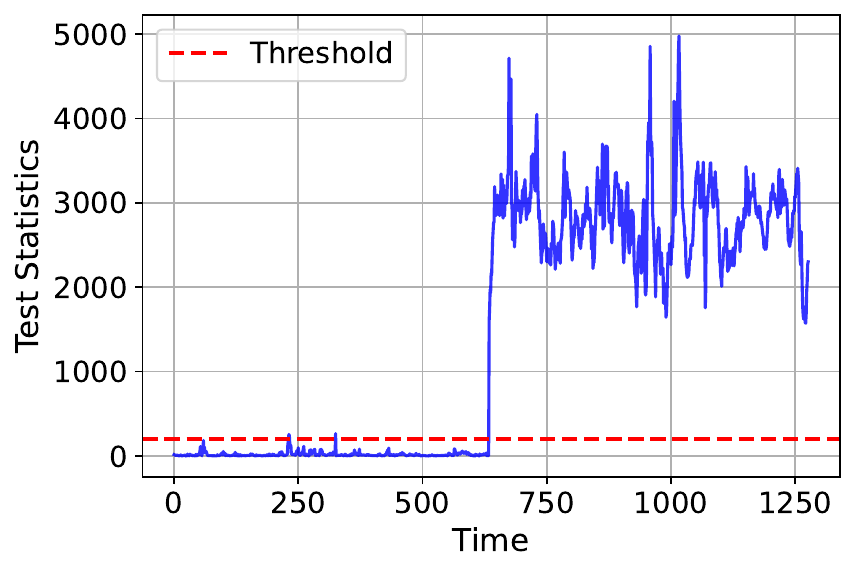}
        \caption{}
    \end{subfigure}
    \caption{ViSTR-GP: Attack 5cm. (a) per-joint residuals (blue) with the MVGP predictive mean and uncertainty band (orange), and (b) frame-wise test statistic with the threshold.}
    \label{fig:5cm}
\end{figure}

Figures~\ref{fig:0.2cm} and~\ref{fig:0.5cm}-\ref{fig:5cm} plot per-joint residuals (blue) together with the MVGP predictive mean and uncertainty band (orange), and the corresponding frame-wise $\chi^2$ test statistic with the threshold. The red dashed line marks the attack onset, separating a normal cycle before and an attack cycle after. Before the attack onset, residuals for all joints are tightly centered within the time-varying MVGP bands, indicating good calibration and low in-control variance. Immediately after onset, systematic drifts emerge from the MVGP predictive distribution, producing sharp spikes in test statistics. Sensitivity scales monotonically with attack severity: even at 0.2 cm the statistic crosses the threshold shortly after onset and does so frequently thereafter; at 0.5 cm–1 cm the exceedance becomes larger and more persistent; at 5 cm the statistic stays above the threshold for long durations. The time-dependent MVGP uncertainty suppresses spurious transients in nominal phases yet tightens around stable segments, which explains the short delays and dense alarms observed across all attack levels.

\begin{figure}[!ht]
    \centering
    \begin{subfigure}[b]{0.7\textwidth}
        \centering
        \includegraphics[width=\linewidth]{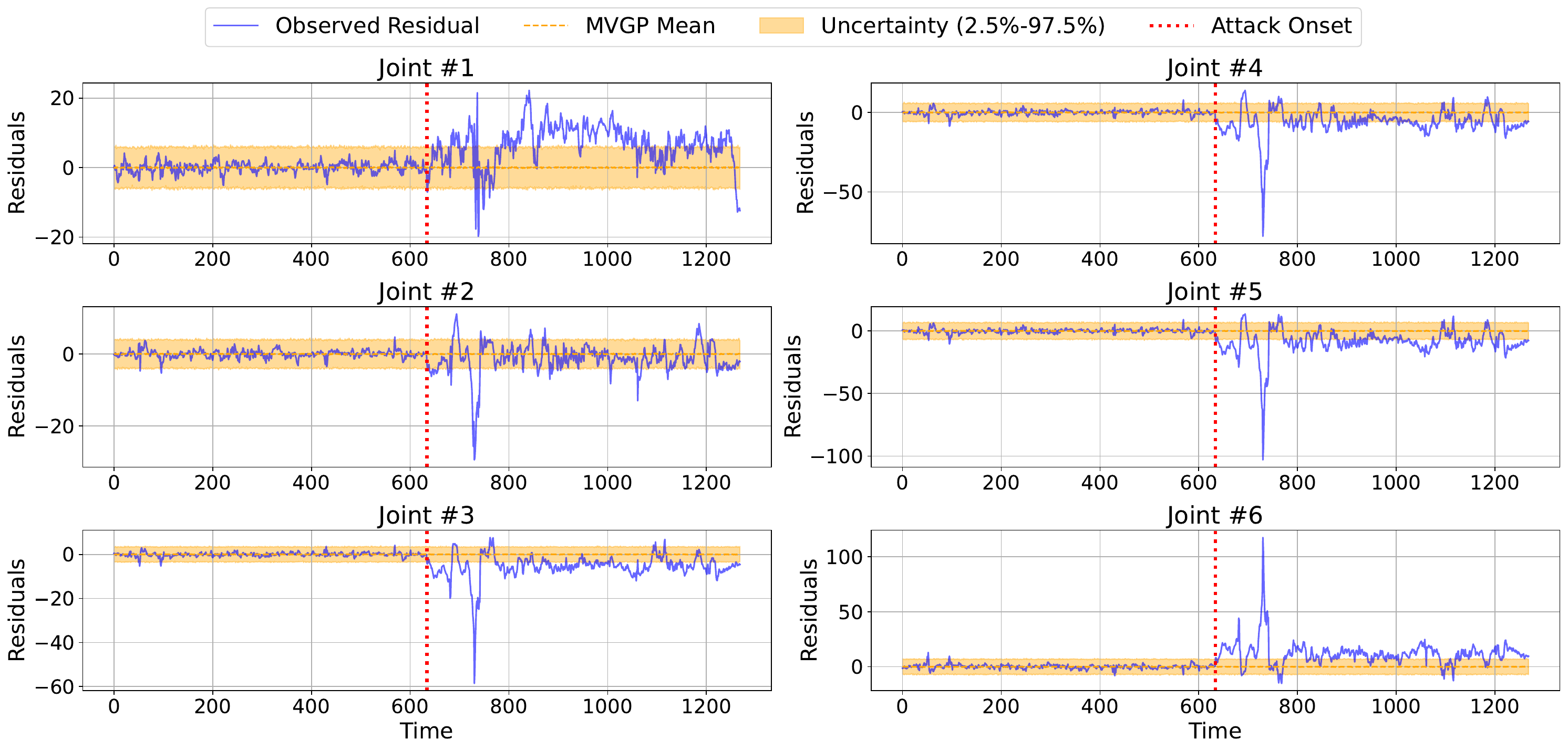}
        \caption{}
    \end{subfigure}%
    \hfill
    \begin{subfigure}[b]{0.3\textwidth}
        \centering
        \includegraphics[width=\linewidth]{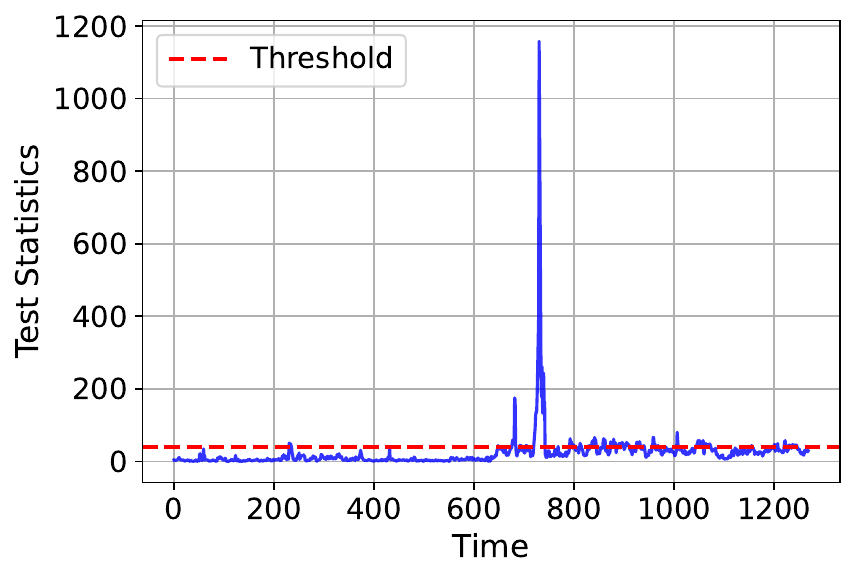}
        \caption{}
    \end{subfigure}
    \caption{TR+IID: Attack 0.5cm. (a) per-joint residuals (blue) with \emph{i.i.d.} mean and uncertainty band (orange), and (b) frame-wise test statistic with the threshold.}
    \label{fig:0.5cm_iid}
\end{figure}

\begin{figure}[!ht]
    \centering
    \begin{subfigure}[b]{0.7\textwidth}
        \centering
        \includegraphics[width=\linewidth]{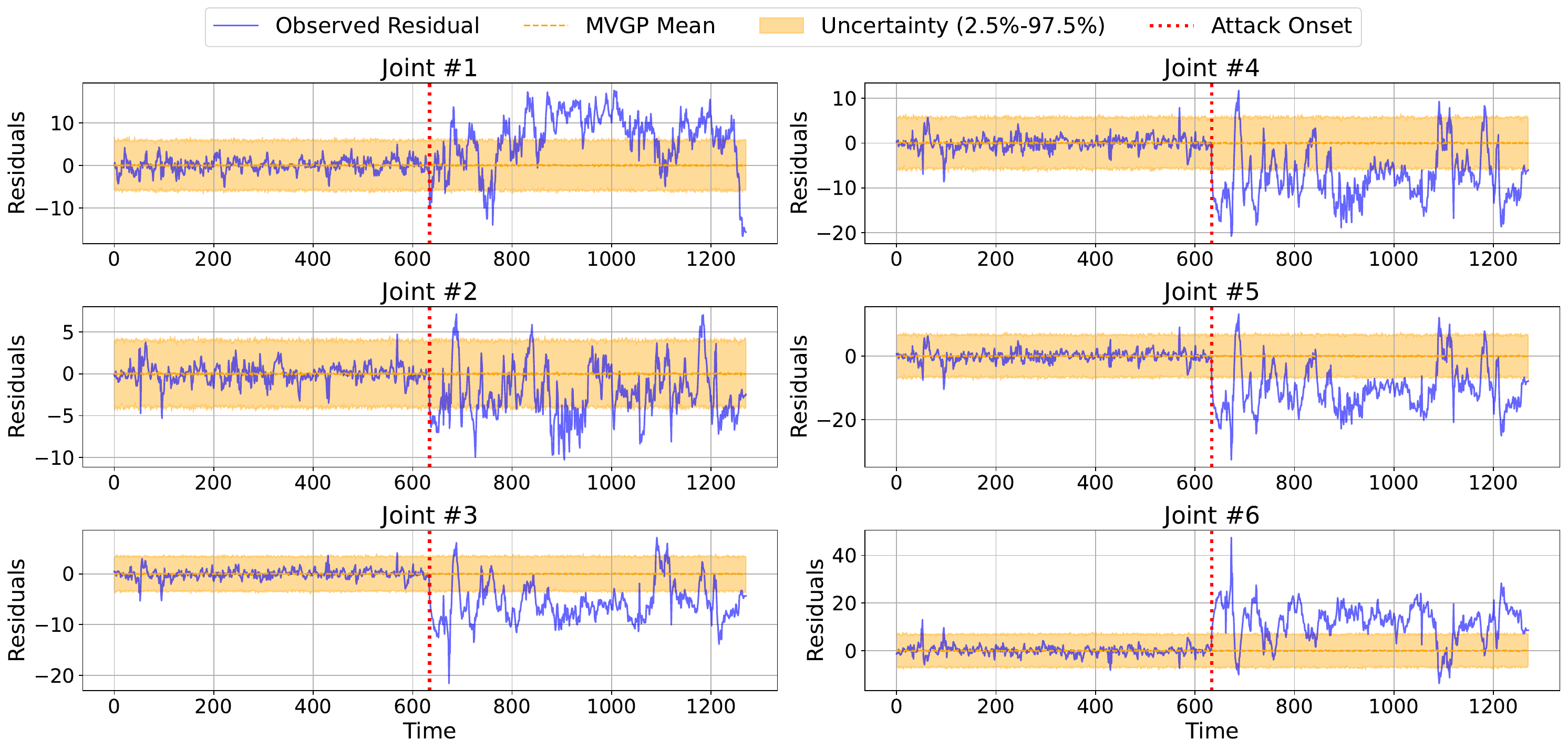}
        \caption{}
    \end{subfigure}%
    \hfill
    \begin{subfigure}[b]{0.3\textwidth}
        \centering
        \includegraphics[width=\linewidth]{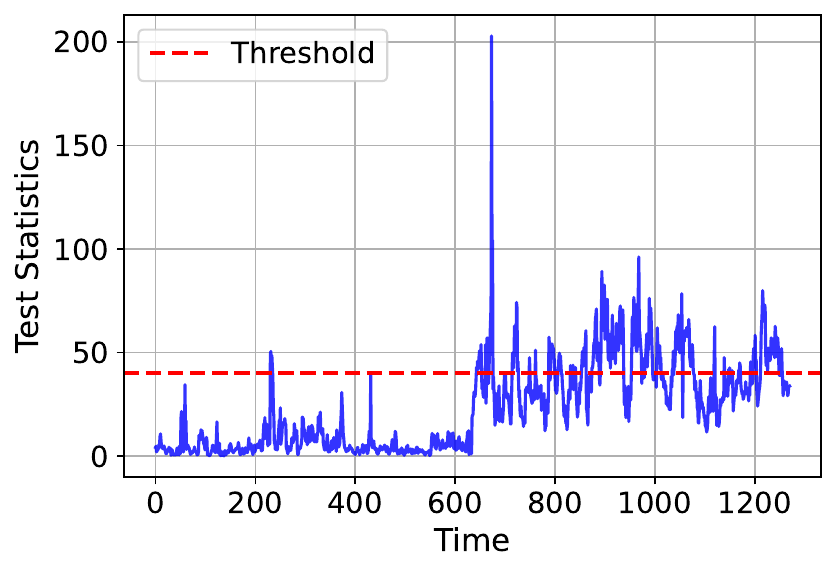}
        \caption{}
    \end{subfigure}
    \caption{TR+IID: Attack 1cm. (a) per-joint residuals (blue) with \emph{i.i.d.} mean and uncertainty band (orange), and (b) frame-wise test statistic with the threshold.}
    \label{fig:1cm_iid}
\end{figure}

\begin{figure}[!ht]
    \centering
    \begin{subfigure}[b]{0.7\textwidth}
        \centering
        \includegraphics[width=\linewidth]{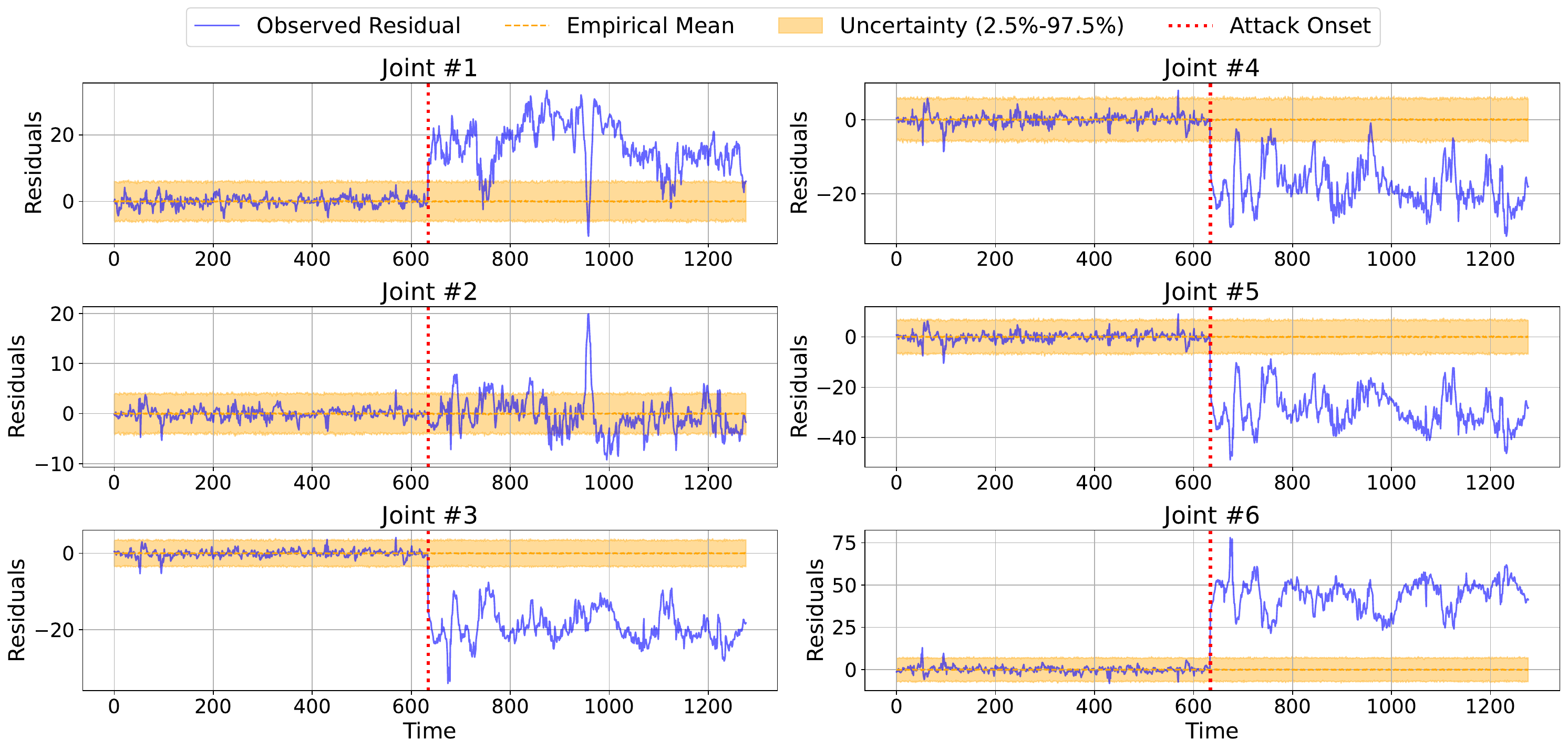}
        \caption{}
    \end{subfigure}%
    \hfill
    \begin{subfigure}[b]{0.3\textwidth}
        \centering
        \includegraphics[width=\linewidth]{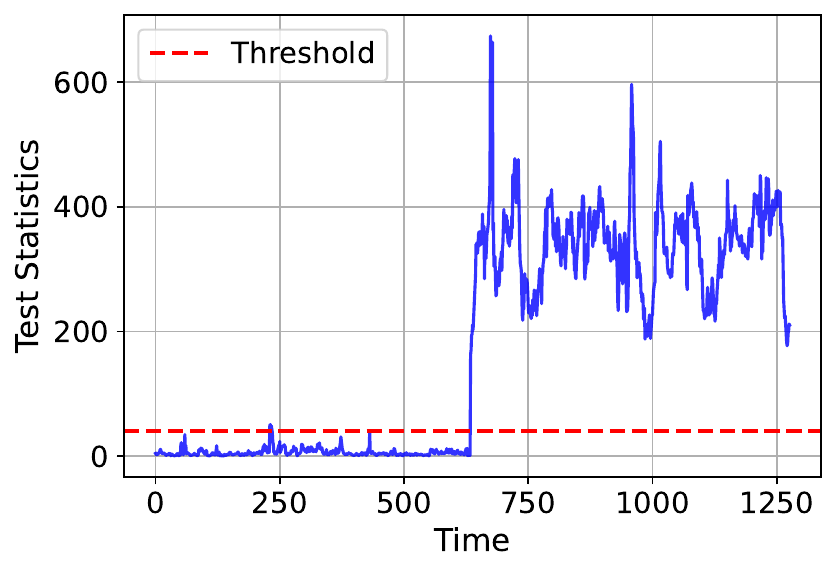}
        \caption{}
    \end{subfigure}
    \caption{TR+IID: Attack 5cm. (a) per-joint residuals (blue) with \emph{i.i.d.} mean and uncertainty band (orange), and (b) frame-wise test statistic with the threshold.}
    \label{fig:5cm_iid}
\end{figure}

Figures~\ref{fig:0.2cm_iid} and~\ref{fig:0.5cm_iid}-\ref{fig:5cm_iid} repeat the analysis when residuals are modeled as \emph{i.i.d.} Gaussian distribution. The empirical mean tracks the pre-attack residuals, but the predictive spread is time-invariant and noticeably wider than in ViSTR-GP. After onset, the residuals shift, yet the inflated, non-adaptive covariance yields slower growth of the $\chi^2$ statistic. At 0.2 cm only intermittent threshold crossings appear; at 0.5 cm–1 cm the exceedance is larger but sparser than with MVGP; at 5 cm sustained alarms occur, though with longer onset delays. Overall, ignoring temporal structure reduces informativeness (wider bands) and lowers alarm frequency, matching the result that TR+IID is less sensitive than ViSTR-GP, especially for subtle attacks.

\end{document}